%% file: arxiv.tex
\documentclass{article} %
\usepackage{iclr2027_conference,times}

\input{math_commands.tex}

\usepackage{hyperref}
\usepackage{url}
\usepackage{xspace}
\usepackage{graphicx}
\usepackage{booktabs}
\usepackage{subcaption}
\usepackage{tikz}
\usepackage{ulem}
\usepackage{multirow}

\newcommand{\mbR}{\mathbb{R}}

\newcommand{\tree}{tree-gram\xspace} %
\newcommand{\trees}{tree-grams\xspace} %
\newcommand{\Tree}{Tree-gram\xspace} %
\newcommand{\Trees}{Tree-grams\xspace} %

\DeclareRobustCommand{\single}{
\tikz[baseline=-0.5ex]{
    \node[circle, draw=black, fill=pink, inner sep=1.5pt] (A) at (0,0) {};
    \node[circle, draw=black, fill=pink, inner sep=1.5pt] (B) at (0.3,0.2) {};
    \draw[blue, ultra thick] (A) -- (B);
}}

\DeclareRobustCommand{\double}{%
\tikz[baseline=-0.5ex]{
    \coordinate (A) at (0,0);
    \coordinate (B) at (0.3,0.2);
    \coordinate (C) at (0.6,0);

    \draw[red, ultra thick] (A) -- ([yshift=-1.2pt]B.south);
    \draw[red, ultra thick] (C) -- ([yshift=-1.2pt]B.south);

    \node[circle, draw=black, fill=pink, inner sep=1.5pt] at (A) {};
    \node[circle, draw=black, fill=pink, inner sep=1.5pt] at (B) {};
    \node[circle, draw=black, fill=pink, inner sep=1.5pt] at (C) {};
}}

\title{What's in a Tree ? Hierarchical $n$-gram models as controlled approximators of language}

\author{%
\makebox[\dimexpr\textwidth-2\tabcolsep\relax][c]{%
\begin{tabular}{@{}c@{}}
Aritra Das\textsuperscript{1,*}\quad
Francesco Cagnetta\textsuperscript{2}\quad
Darshil Doshi\textsuperscript{3}\\[0.5em]
Victor V. Albert\textsuperscript{1}\quad
Maissam Barkeshli\textsuperscript{4,5}\quad
Andrey Gromov\textsuperscript{6}
\end{tabular}%
}%
}

\arxivcopy
\begin{document}

\maketitle

\begingroup
\renewcommand{\thefootnote}{*}
\footnotetext[7]{Corresponding author. Email : \textit{aritrad@umd.edu}}
\endgroup
\footnotetext[1]{Joint Center for Quantum Information and Computer Science, UMD \& NIST.}
\footnotetext[2]{Simplex.}
\footnotetext[3]{Data Science and AI Institute, and Department of Physics and Astronomy, JHU.}
\footnotetext[4]{Department of Physics, UMD.}
\footnotetext[5]{Joint Quantum Institute, UMD.}
\footnotetext[6]{Independent.}
\begingroup
\renewcommand{\thefootnote}{$\dagger$}
\footnotetext[8]{The code is released on  \href{https://github.com/a-writes-code/treegram.git}{Github}. Saved models are released on \href{https://huggingface.co/hugs-from-a-25/tree-gram-generators}{Hugging Face}, while generated datasets are released as  \href{https://huggingface.co/datasets/hugs-from-a-25/tree-gram-text}{Hugging Face datasets}.}
\endgroup

\setcounter{footnote}{6}

\maketitle

\begin{abstract}
Hierarchical models have been proposed as synthetic settings for studying language. However, their relevance to realistic language corpora remains underexplored.
In this work, we introduce \trees, autoregressive generative $n$-gram models that build the conditional probability tensor $p(x_n \mid x_{1:n-1})$ through a hierarchical composition of order-2 and order-3 tensors.
$n$ controls the maximal context window, whereas a rank hyperparameter $d$ controls model capacity, yielding an interpretable and systematically scalable family of approximators of language.
We show how \trees outperform existing $n$-gram baselines, while remaining competitive with deep transformer-based language models.
We demonstrate the utility of such controlled approximations by using \trees as generative models. 
In particular, we show that hyperparameter transfer under $\mu$P can be made more robust when narrower models are tuned on simpler data, generated by \trees with moderate $n$ and $d$.
In addition, we show that entropy and effective range of correlations control performance gap after fine-tuning. This is because \trees allow us to explicitly control horizon and capacity. 
When allowed to route across candidate topologies, we observe emergent sparsity in \trees. 
Such concentrated routing permits pruning upto 95\% of topologies without significant change in performance.
\end{abstract}

\section{Introduction}

Language is hypothesized to be hierarchically structured \citep{Chomsky1956,Chomsky1957,Everaert2015,Hauser2002,Goodfellow2016}.
Its statistical structure has been studied using classical and modern language models, including Markov-chain and $n$-gram models \citep{Shannon1951,Brown1992,Bengio2003}. 
Synthetic datasets inspired by context-free gramars~\citep{Chomsky1956,HopcroftUllman1979,KleinManning2004,Kim2019} have also been proposed as tractable testbeds to understand language modeling. 
In particular, \citep{PhysRevX.14.031001} propose a family of tree-like generative models where a hierarchy of latent nodes determines the probability distribution of the leaves. 
Tree-like generative models can successfully represent compositional dependencies and improve language modeling~\citep{Dyer2016,Kuncoro2018}; however, learning linguistically meaningful tree structures directly from data remains difficult \citep{KleinManning2004,Kim2019,HewittManning2019}. It is also important to understand what tree-like properties are present in real data, which can be accomplished by creating trainable versions of such models.

In this work, we propose such a trainable model. In particular, we propose \textbf{\trees}, which are hybrid $n$-gram models based on hierarchy and local connectivity. \Trees combine order-2/order-3 tensor approximation with normalization layers, residual connections, and MLPs to approximate the order-$n$ transition tensor $p(x_n| x_{1:n-1})$. The two hyper-parameters $n$ and $d$ are independent controls on \trees, while $n$ controls the \textit{horizon} of the model, $d$ controls model \textit{capacity}. A summary of our contributions is provided below.

\paragraph{Summary of contributions} %

\begin{itemize}
    \item We present \trees as generative models for controlled approximation of language. The approximation is controlled through a horizon parameter $n$ and a rank parameter $d$.
    In contrast with context-free grammars, \trees learn the latent features from data.
    We train \trees on TinyStories~\citep{eldan2023tinystoriessmalllanguagemodels} as a proof of concept, for which we demonstrate how \trees out-perform existing $n$-gram estimators and remain competitive with deep neural references.
    \item We demonstrate the utility of such controlled approximations by analyzing hyperparameter (HP) transfer and fine-tuning performance. 
    In particular, we show that HP transfer is more robust when the smaller proxy is trained on lower-complexity synthetic data.
    We also investigate the effect of alignment between pre-training and finetuning data, where we show that performance has a stronger dependence on the finetuning distribution.
    \item We observe concentrated topology-routing distributions when \trees are allowed to choose topologies at each layer through linear routers. We observe that the predictive performance changes very little even when we prune upto 95 \% of topologies at each level.
\end{itemize}
We emphasize that the value of tree-grams is \textit{experimental}. It allows us to rigorously test ideas about the structure of (language) data and learning. This is in contrast with \textit{empirical} value as a model class that out-performs SOTA neural networks. Thus the focus in this work was to set up a basic version of tree-like models that scale better than existing $n$-gram baselines.

\begin{figure}[t!]
\centering
\includegraphics[width=\textwidth]{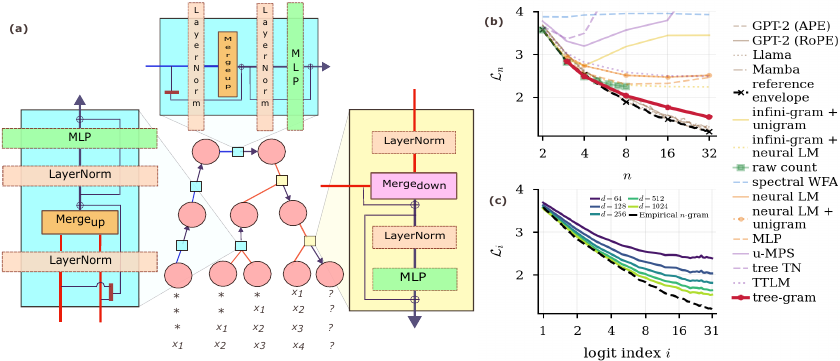}
\caption{\textbf{(a) The \tree architecture.} 
\Trees consist of hybrid blocks which are used to merge intermediate nodes, progressively reducing the number of nodes $n^{(\ell)}$ at each layer.
Red edges are order-$3$ merges (\double), blue edges denote order-$2$ merges (\single).
All upward merges (denoted with an upward-pointing arrow) are followed by LayerNorm and an MLP, shared across the first $n^{(\ell)}-1$ nodes of the layer.
Autoregressive prediction is done by masking ($*$) previous tokens.
The last node outputs logits for next-token prediction.
\textbf{(b) Scaling with the horizon $n$.} cross-entropy of the next-token probability given exactly $n\,{-}\,1$ tokens of context. \Trees (in red) outperform tested $n$-gram estimators for large $n$ but fall short of deep transformers.
\textbf{(c) Scaling with the rank $d$.} Per-token loss as a function of the rank $d$. We observe a smooth approach towards the empirical lower envelope of the available references as $d$ increases.
The corresponding plots for Wiki-Text and FineWebEdu are presented in the Appendix~\ref{app:extra}.
}
\label{fig:comparator}
\label{fig:scaling}
\label{fig:intro}
\end{figure}

\subsection{Background and Motivation}
\label{sec:related}

Hierarchical models have been used to study synthetic datasets resembling language. \citep{Chomsky1956,ChomskySchutzenberger1963,yao2021selfattention,PhysRevX.14.031001}.
Justification for such hierarchical modelling comes from statistical property analysis \citep{thurner2015understanding}, information theory \citep{futrell2026linguistic, yang2026functionwordsstatisticalcues} and
cognitive representations \citep{Crain_Nakayama_1987, doi:10.1073/pnas.1018711108, ding2016cortical, hale-etal-2018-finding}. 
However, all such models fix hidden features based on semantic or syntactic structures. 
\Trees are introduced as generative models where the hidden node features are directly learnt from data.
In addition, it is useful to have independent knobs that control the complexity of the full corpus.
Such an approximation can be made by either controlling the capacity or the horizon of the generative model \citep{cagnetta2024scaling}.
Such approximations have been made by using traditional and smoothed $n$-gram counts \citep{chen1996smoothing, parikh2014plre, liu2025infinigramscalingunboundedngram},  hidden Markov models \citep{balle2014spectral} or tensor-trains \citep{miller2021tensor,su2024tensortrain}.
However, these methods are difficult to scale beyond few words.
On the other hand, the effect of various architectural hyper-parameters in transformers on model capacity is poorly understood.
We introduce \trees in light of the above considerations.
Detailed background and related works are presented in Appendix~\ref{app:related}.

\section{The \tree architecture}
\label{sec:arch}

We implement \trees as a hierarchical architecture (see Fig.~\ref{fig:intro}), with layers indexed by $\ell$.
After an initial shared token embedding $\operatorname{Emb}$,
\begin{equation}
    \label{eq:emb}
    z^{(0)}_i = \operatorname{Emb}(x_i)+p_i \in \mbR^{d},
\end{equation}
the new layer is formed via local order-2 ( \single ) and order-3 (\double ) merges.
$p_i$ is a (learnable) positional encoding.
The $\operatorname{Merge}$ operation is defined as
\begin{equation}
    \label{eq:merge-up}
    h^{(\ell)}_i = \operatorname{Merge}_{\text{up}}(z) = 
    \begin{cases}
    W^{(\ell-1)}_i z^{(\ell-1)}_i & (\single \text{ connection}), \\[4pt]
    W^{(\ell-1)}_i 
    \left(z^{(\ell-1)}_i \odot z^{(\ell-1)}_{i\,{+}\,1}\right) & (\double \text{ connection}),
    \end{cases}
\end{equation}
where $z^{(\ell)}_j \in \mbR^{d}$, $W^{(\ell-1)}_j \in \mbR^{d\times d}$, and $\odot$ denotes elementwise multiplication. The order-3 merge is a CP-type (Hadamard) factorisation of an order-3 tensor, as in the tensor circuits introduced by \citep{loconte2025relationshiptensorfactorizationscircuits}. %
The factorization above is a non-unique choice. It restricts the class of topologies, although the number of topologies is super-polynomial $\mathcal{O}(n^{\sqrt{n}})$. Moreover, it fixes the depth to be $\log n$ , which we call the principle of \textit{minimal depth}. \footnote{The name is inspired from the work of \cite{ganardi2020balancingstraightlineprograms} who show that a context-free grammar of size $m$ and string length $n$ can be converted into a balanced binary tree with depth $\mathcal{O}(\log n)$ in linear time.}
\Trees belong to a more general class of hierarchical tensor decompositions with latent nodes \cite{pmlr-v28-song13,razin2022implicitregularizationhierarchicaltensor, furman2026benignlosslandscapescoexist}. 
\textit{Any} factorization of the nodes represents a valid \tree architecture.

Given the horizon, or maximal context window, $n$, the number of nodes in each layer is defined via a recursive relation
\begin{equation}
    \label{eq:nodes}
    n_{(\ell)} = \left\lfloor n_{(\ell-1)}/2\right\rfloor + 1; \quad n_{(0)} \equiv n,
\end{equation}
fixing the depth of the network is $\lceil\log_2(n-1)\rceil$.
For each layer, $n_{(\ell)}$ and $n_{(\ell+1)}$ constrain the total number of order-3 merges, but their position along the sequence can vary.
The choice of positions at each layer determines the \textit{topology} $\tau$ of the \tree.

The merge operation is preceded by LayerNorm, and followed by the residual+MLP composition typical of transformer layers,
\begin{equation}
\begin{aligned}
\bar z^{(\ell-1)}_i&=\operatorname{LN}_{\ell,\text{proj}}(z^{(\ell-1)}_i),\quad
q^{(\ell-1)}_i= \operatorname{Merge}\left(\bar z^{(\ell-1)}_i\right)
+\frac{\rho}{2}z^{\text{res},\ell-1}_i,\\
z^{\ell}_i&=\rho q^{(\ell-1)}_i+\operatorname{MLP}_{\ell}(\operatorname{LN}_{\ell,\text{mlp}}(q^{(\ell-1)}_i)).
\end{aligned}
\label{eq:composite-node}
\end{equation}
where $z^{\text{res},\ell-1}_i = \bar{z}_i$ if $i$ has a single connection, and $z^{\text{res},\ell-1}_i = \bar{z}_{i} + \bar{z}_{i\,{+}\,1}$ if it has a double connection.
The LayerNorm ($\operatorname{LN}$) and MLP are shared across nodes in a layer, while the \tree weight depends on the parent position. Unless stated otherwise, $\mu=1$, the MLP uses GELU and hidden width $4d$, and all reported \tree runs use learned positional embeddings to break the commutativity of the \double merge. 

The downward pass occurs along the terminal nodes at each layer, with the same residual/MLP pattern as Eq.~\ref{eq:composite-node}. The only difference is that $\operatorname{Merge}_{\text{down}}$ operation is modified as 
\begin{equation}
    \label{eq:merge}
    h^{(\ell)}_i = \operatorname{Merge}_{\text{down}}(z) = 
    \begin{cases}
    (W^{(\ell+1)}_i)^T z^{(\ell+1)}_{c^\ell_i} & (\single \text{ connection}), \\[4pt]
    \left(z^{(\ell+1)}_{c^\ell_{i,L}} \odot (W^{(\ell+1)}_i)^T z^{(\ell+1)}_{c^\ell_{i,R}}\right) & (\double \text{ connection}),.
    \end{cases}
\end{equation}
The terminal node in the bottom layer serves as the predictive node, where an embedding-weight-tied softmax head maps it to next-token probabilities.
\Trees perform autoregressive prediction by unfolding each sequence into its $n\,{-}\,1$ left-padded prefixes, which activate different subtrees. 
The model was trained with the usual autoregressive cross-entropy loss.
Appendix~\ref{app:architecture} provides further details on the architecture.

For the primary experiments reported in the main text, we train on a 529M-token subset of TinyStories, a synthetic language corpus \citep{eldan2023tinystoriessmalllanguagemodels}, using a custom BPE tokenizer with vocabulary size $V=8192$ \citep{cagnetta2024scaling}. Details on optimization are in Appendix~\ref{app:exp}.
We train on two additional datasets : (1) a 130MT WikiText-103 \cite{merity2017pointer} dataset, obtained by designing a custom BPE tokenizer with vocabulary size $V=8192$  and (2) a 1.05BT FineWeb-Edu \cite{penedo2024the} dataset, tokenized with the GPT-2 tokenizer and vocabulary size $V=50,257$.

\section{Results}

\subsection{Comparison to existing baselines}
\label{sec:baselines}

\begin{figure}[b!]
\centering
\includegraphics[width=\textwidth]{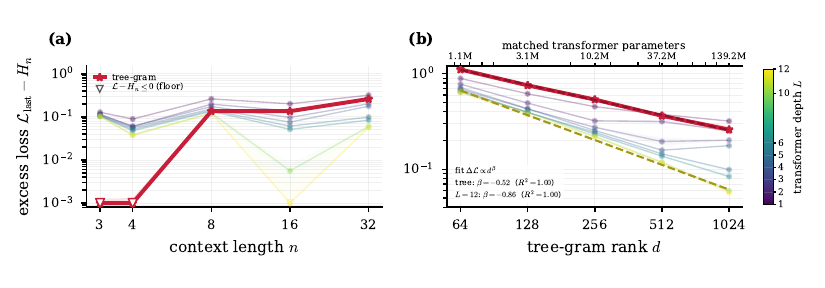}
\caption{Last-token loss of a fixed \tree (red) against matched transformers of varying depth $L$, matched to the \tree in parameter count, context and optimization steps. 
\textbf{(left)} Scaling with horizon $n$ at $d=1024$: \trees perform better at small $n$, but are outperformed by the deeper transformers at longer contexts. 
\textbf{(right)} Scaling with \tree rank $d$ at $n=32$. 
Transformers plateau earlier at smaller depth. 
\Tree loss continues to decrease with $d$ over the tested range, while shallow transformers plateau. Thus \trees appear to be scalable.
$H_n$ is the (empirical) n-gram entropy.
}
\label{fig:treevstfr}
\end{figure}

Figure~\ref{fig:intro} b) compares \trees to different classes of $n$-gram estimators, from raw counts to transformer-based LLMs.
In particular, we consider classical smoothed counting models \textbf{\textit{(raw count)}} \citep{chen1996smoothing,parikh2014powerlowrank}, unbounded suffix array calculations \textbf{\textit{(infinigram)}} \citep{liu2025infinigramscalingunboundedngram}, weighted finite-automaton approximations obtained via spectral methods \textbf{\textit{(spectral WFA)}} \citep{balle2014spectral}, tensor train approximations \textbf{\textit{(tree TN, u-MPS, TTLM)}} \citep{miller2021tensor, su2024tensortrain}, MLP-based neural models \textbf{\textit{(MLP, neural LM, neural LM + unigram)}} \citep{MLPEmulation, bengio2003neural, blat2005hybrid}, transformer-based \textbf{\textit{(GPT-2 (APE), GPT-2 (RoPE), Llama)}} and Selective-State-Space-Models-based \textbf{\textit{( Mamba)}} language models.
A detailed description of these baselines is found in Appendix~\ref{app:baselines}.

In the range of horizons probed, $n\,{\leq}\,32$, the LLM baselines converge to the model-independent asymptotic conditional entropy of the true next-token distribution~\citep{cagnetta2024scaling} (black dashed line in Figure~\ref{fig:intro}b and c). 
The model-independence is a strong indication that this is close to the true $n$-gram entropy. \citep{TfrNGramStats,krishnan2023ngramapproximationpretrainedlanguage,cagnetta2024scaling}
This will be referred to as the empirical entropy $H_n$ in what follows.
\Trees match the empirical $n$-gram reference up to $n\,{\approx}\,8$, vastly outperforming all other baselines except LLMs (Figure~\ref{fig:intro}b). 
Furthermore, increasing the capacity parameter $d$ reduces the performance gap to LLMs (Figure~\ref{fig:intro}c)

\subsection{Comparison with transformers}

In order to understand the performance gap of \trees to LLMs more carefully, we compare \trees with transformers of different depth with the same \textit{parameter count}, the same \textit{context length} and trained with the same \textit{number of optimization steps}, thereby highlighting the difference in their architectural biases.  
For each \tree target, we take a  decoder-only transformer of depth $L \in \{1,2,3,4,5,6,10,12\}$. 
\Trees perform better than the shallow matched transformers. This pattern holds until approximately $n=8$.
At longer contexts, the \tree is unable to match deep transformers. 
The right panel shows that the excess loss monotonically decreases with $d$. 
While shallow transformers plateau quickly, \trees monotonically decrease with a power-law curve, showing that they are more scalable than shallow transformers. 
In addition, the slope is not drastically different from the deepest transformers tested.
It is worth emphasizing that this performance gap could simply be due to sub-optimal optimization routines for \trees. 
Developing better principles for optimizing \trees is left for future work.
The per-position curves behind these endpoints are in Fig.~\ref{fig:treevstfrgrid} (Appendix~\ref{app:treevstfr}), where the \tree tracks the transformer family for the first few positions and drifts away at long context, the same signature as in Fig.~\ref{fig:comparator}.

\section{\Trees as Controlled Approximation of Language}

We use the \trees as a source of synthetic data, with $n$ and $d$ independently controlling distinct modes of approximation of the synthetic corpus to the original training corpus.
We emphasize that, unlike transformers, a single hyper-parameter $d$ controls the model capacity.
In Appendix~\ref{app:gen_data} we present a qualitative analysis of the decoded text generated by \trees from the same prompt. 
Increasing $n$ (at sufficiently large $d$) increases coherence from word-level ($n=3$) to sentence-level ($n=32$). 
On the other hand, increasing $d$ increases both coherence \textit{and} quality of the generated text.
We demonstrate the scientific utility of such finite-window and finite-capacity approximations by experimenting on hyper-parameter transfer under $\mu$P and effect of pre-training and fine-tuning distribution shifts on fine-tuning performance.

\subsection{Hyperparameter transfer under Maximal Update Parametrization }

The standard recipe for hyperparameter (HP) transfer under the $\mu$P (\underline{M}aximal \underline{U}pdate) parametrization \cite{yang2022tensorprogramsvtuning} involves two stages: \textit{tuning} and \textit{transfer}. The tuning stage involves training a small model, initialized according to $\mu$P to find the optimal HP combination
This optimal HPs are then transferred to a larger model through appropriate scaling.
However, the complexity of the datasets in the tuning phase is only controlled implicitly, for example through the amount of pre-training data \cite{bjorck2025scalingoptimallrtoken, li2025predictablescaleistep}, data deduplication \cite{lee-etal-2022-deduplicating, abbas2023semdedupdataefficientlearningwebscale} or high-quality data pruning \cite{sorscher2023neuralscalinglawsbeating}.

\begin{figure}[t!]
    \centering
    \includegraphics[width=\linewidth]{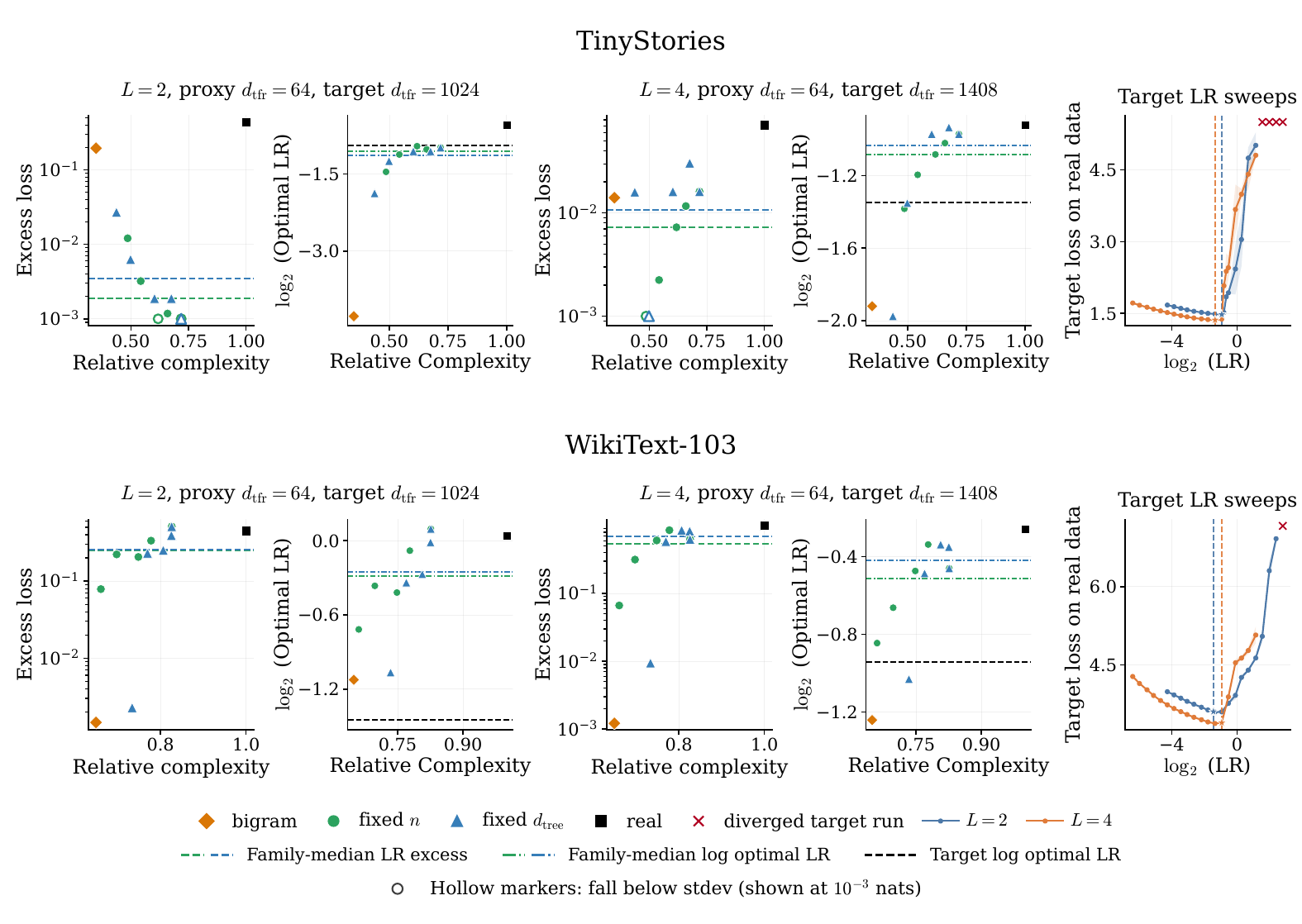}
    \caption{Hyper-parameter (HP) transfer from narrow proxies trained on controlled synthetic approximations of the target corpus. 
    The excess loss is the difference between the loss of a wide model ($d{=}1024$ at $L{=}2$ and $d{=}1408$ at $L{=}4$) in at the proxy's optimal learning rate $\eta^\star$ and the wide model's own HP-tuned minimum.
    Tuning on different distributions gives different optimal learning rate.
    Almost all tuning with simpler proxy data gives a lower excess loss than if tuned on real data.
    In addition, we see a \textit{monotonic} drift of the optimal learning rate with data complexity.
    The drift explains the excess loss because of the asymmetric loss curves around the optimal learning rate. %
    Relative complexity is defined as $H_{\text{real}}/H$, where $H$ stands for the entropy of a particular distribution.
    }
    \label{fig:gendata}
    \label{fig:hpt}
\end{figure}

Using \trees as a generative model, we can control the data complexity explicitly. 
We observe that tuning HPs on a controlled-complexity approximation of the original dataset yields better transfer.
Specifically, we tune the learning rate of 2- and 4-layer $\mu$P decoder-only proxy transformers with widths $d\in\{64,\dots,256\}$, then analyze the transfer to target transformers with various widths.
We judge the transfer by the excess loss of the target model with transferred learning rate with respect to its own HP-tuned minimum.
As shown in Figure~\ref{fig:gendata}, tuning $\nu^\star$ on the original data almost always yields the highest excess loss (black square).
The converged loss has an asymmetric dependence around the optimal learning \citet{kosson2026weight,ishikawa2026measuringlocalshapelearningrate,filatov2025time}. 
Thus, a drift in the optimal learning rate can lead to a higher excess loss. We show empirical evidence for this drift in Figure~\ref{fig:gendata}.
The  $\nu^\star$ from tuning on the real dataset however shows a non-monotonic effect with width, as shown by the black dashed line which sits lower than the optimal learning rate obtained from tuning on the real data.
While a formal characterization of which proxy distribution to tune on, remains an open question - choosing the median learning rate over different proxy distributions can give robust transfer.
It is worth emphasizing that the undershooting the learning rate is better only if the optimal learning rate at a narrower width is \textit{higher} than that at larger width. 
In the Appendix, we show that this does \textit{not} hold for FineWeb-Edu. The monotonic drift in optimal learning rate as a function of data complexity however still holds, which is the relevant phenomena.

A complete mechanistic understanding of this observation is beyond the scope of the present work.
However, we present a simple toy model of data from \trees, which we show can give rise to a mechanism for the drift in the optimal learning rate $\eta^\star$. To keep notation simple, we will work in log-space $\nu^\star \equiv \log_2 \eta^\star$. 
The model is a Gaussian surrogate of hierarchical features, in the spirit of the Random Hierarchy Model~\citep{PhysRevX.14.031001} and of multiscale Gaussian signal models~\citep{wornell1992estimation}.
We start with a left-branching binary tree over the $n-1$ context tokens. The node at level $k$ summarizes a block of $2^k$ tokens, so there is one node per level, $k=0,\ldots,K$. 
This sets the number of levels to be $K \equiv \lfloor\log_2[n - 1]\rfloor$. The power at level $k$ is $\pi_k = r^k/Z_K$ for some constant $r<1$, where $Z_K=\sum_{\ell\le K}r^\ell$ fixes the total power to one for every $n$.
The importance of a node is controlled by the variance of its coefficient, $S_{k,a} \sim \mathcal{N}(0, \pi_k/d)$. We assume that the power of each node is diluted evenly over $d$ coordinates. If we define unit-norm vectors $u_{k,a} \in \mathbb R^V$, then the next-token logits are
\begin{align}
g(S)=\sum_{k=0}^{K(n)}\sum_{a=1}^{d}S_{k,a}\,u_{k,a}\in\mathbb R^V,
\\
Y\mid S\sim\operatorname{Categorical}\big(\operatorname{softmax}(g(S))\big).
\label{eq:toy-label}
\end{align}
We also assume that the observed features are noisy versions of the true features $$\Phi_{k,a}=\sqrt q\,S_{k,a}+\sqrt{1-q}\,\Xi_{k,a}$$ 
where the noise term 
$\Xi_{k,a}\sim\mathcal N(0,\pi_k/d)$ is independent of $(S,Y)$.
The quality factor $q(d)=\sqrt{d}/(\sqrt{d}+\sqrt{d_0})$
sets the signal-to-noise ratio $q(d)/(1-q(d))=\sqrt{d/d_0}$.

Solving a minimal quadratic model like that of \citep{minimal-quadratic-model} with aligned features, the single-step optimal learning rate has the following dependence:
\begin{equation}
    \label{eq:nu-star-toy}
    \nu^\star_{\Lambda} = 
    \underbrace{\nu^\star_{\infty}}_{\text{asymptote}} +
    \underbrace{\Delta(\Lambda)}_{\text{data complexity}} 
\end{equation}
Here $\Lambda \equiv \Lambda(n,d)$ is a quantity that characterizes the complexity of the synthetic data. 
$\Delta(\Lambda)$ is a term that depends solely on the approximation parameters $n$ and $d$.
$\Delta(\Lambda)$ increases monotonically with $d$ and $n$, which provides a mechanism for a drift in the optimal learning rate. %
The full derivation is presented in Appendix~\ref{app:dyadic-toy}.
We emphasize that the toy model presented is not intended to provide a causal mechanism for our empirical observations.
\begin{figure}
    \centering
    \includegraphics[width=\linewidth]{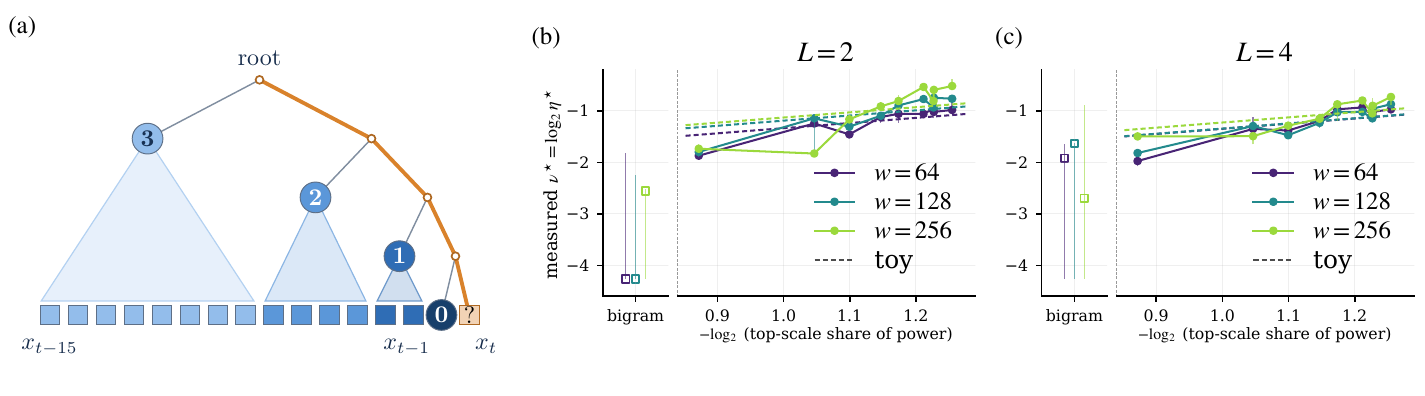}
    \caption{
    \textbf{(a)} The toy model for \tree data.
    \textbf{(b), (c)} We show that the dependence of the optimal learning rate on the measured allocated power share matches closely with the prediction from the toy model.
    Note that the bigram model is not a \tree and hence excluded from the fit.
    }
\end{figure}

\subsection{Pre-training and finetuning}

The value of a pre-training distribution depends on how well it aligns with the fine-tuning distribution.
\citep{isik2025scaling} find that downstream quality improves monotonically with more pre-training data only when the two distributions are well aligned.
Synthetic corpora have been used to ask which structure transfers to natural language: music and code \citep{papadimitriou2020learning}, artificial languages with controlled dependency structure \citep{ri2022pretraining,chiang2022transferability}, and formal languages with hierarchical dependencies \citep{hu2025circuits} can all transfer, depending on the structure they carry.
\Trees provide a structured manner to introduce, and to measure, misalignment between the pre-training and finetuning data, through either the context horizon $n$ or the model capacity $d$.

We clarify that by fine-tuning we mean continued next-token training of all parameters on the target corpus, as in studies of transfer between text distributions \cite{hernandez2021scaling, hu2025circuits}. Unlike a narrower downstream task, a target here can be more complex than its source, so we avoid low-rank adapters \cite{biderman2024lora}.
We re-warm and cosine re-decay the learning rate from a fresh optimizer state \cite{ibrahim2024simple, gupta2023rewarm}, but without the replay that \cite{ibrahim2024simple} recommend. This is because replaying source data would mix the source into every target run and remove the same-distribution control on the diagonal.
For a generator $q$, we estimate its Bayes floor on the sampled path as
\begin{equation}
H_q = \frac{1}{N}\sum_{t=1}^{N}\left[-\sum_{v\in\mathcal V}q(v\mid x_{t-n+1:t-1})\log q(v\mid x_{t-n+1:t-1})\right].
\label{eq:bayes-floor}
\end{equation}
The conditional entropy at each sampled context is computed from the generator's full softmax during generation --- it is not inferred from the downstream model's validation loss. Appendix~\ref{app:ptft} specifies the model ladder, data split and optimization schedule.%

\begin{figure}
    \centering
    \includegraphics[width=\linewidth]{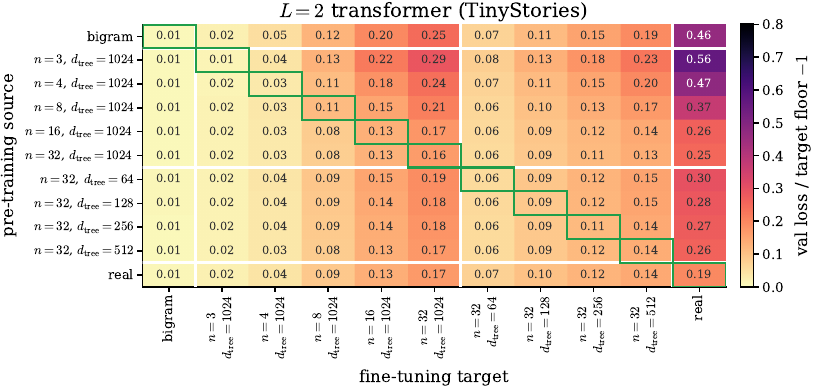}
    \caption{Finetuning on controlled distributions after pre-training. 
    We plot the relative difference between the final validation loss achieved after finetuning and the target Bayes floor of the target distribution.
    There is a monotonic worsening of performance as  the fine-tuning target complexity increases. 
    In addition, the performance monotonically worsens if we \textit{pre-train on too simple of a data}.
    For brevity, we display the results of a 2-layer transformer only for TinyStories. The other plots are included in Appendix~\ref{app:extraptft}.
    }
    \label{fig:transfer-mt}
    \label{fig:ptft}
\end{figure}

Figure~\ref{fig:ptft} reports the result of this experiment. 
The final finetuning performance depends monotonically on the complexity of the finetuning distribution. 
The dependence on the pre-training distribution is more nuanced. If the pre-training data is \textit{too simple}, we notice a monotonic worsening of performance with decreasing complexity.
However, when fine-tuning distributions are simple, finetuning performance is almost independent of the pre-training distribution.
Note that we separate the capacity-limited and horizon-limited modes Figure~\ref{fig:ptft}.
This is because the monotonicity is a function of the entropy of the distribution \textit{within the same approximation mode} (the figure sorted by entropy alone is reported in Appendix~\ref{app:extra}, where the clean trend shown disappears.)
Most of the gap is set by the target, through two properties that can be measured directly from its token statistics: its entropy rate and the effective range of correlations,
measured by how quickly the mutual information between tokens decays with their separation, beyond the horizon. Entropy has the larger effect, but at \textit{comparable} entropy a target
with a longer effective range is consistently \textit{harder} to reach. 
The joint effect of entropy and effective range of correlations was discussed in \cite{cagnetta2024scaling}.
We propose an \textit{ad hoc} four-parameter law in these two quantities, plus a penalty when the source carries less information than the target, which accounts for 82--99\% of the variance across all source--target pairs
in every setting, and predicts the gap for held-out targets. Details are delegated to Appendix~\ref{app:ptft-law}.

\section{Ablations}
\label{sec:ablation}

\paragraph{Role of topologies} 
 
\begin{figure}[t!]
    \centering
    \includegraphics[width=\linewidth]{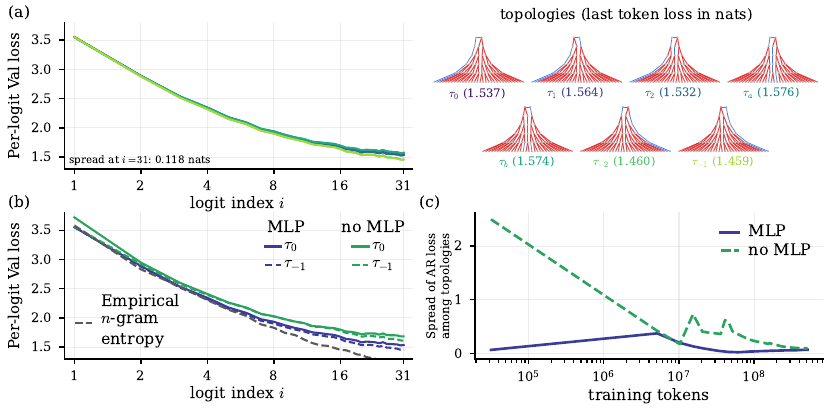}
    \caption{
    \textbf{(a)} Role of topologies. (Left) At convergence, the topologies show slight difference at long $n$. (Right) The seven different topologies we experimented on. The anti-aligned topologies seem to significantly out-perform other topologies only for last-token prediction (losses in nats mentioned).
    \textbf{(b)} Role of MLPs. MLPs have a significant effect in reducing the loss even at matched parameter count.
    \textbf{(c)} MLPs help control the drift in the losses for different topologies, giving more stable training. 
    }
    \label{fig:ablation}
\end{figure}

We focus on three kinds of topologies: aligned, balanced, and anti-aligned. 
\footnote{This naming convention comes from the hypothesis that English language is predominantly right-branching \cite{BERG_2012}. We emphasize that we do not test the validity of this assumption.}
The aligned topologies contain all \single connections on the left-hand side (topologies labelled $\tau_0$, $\tau_1, \tau_2$ in Figure~\ref{fig:ablation}), while the anti-aligned topologies contain all \single connections on the right-hand side (labelled $\tau_{-1}$, $\tau_{-2}$). Balanced topologies have the \single connections in the middle (labelled $\tau_a$, $\tau_b$).
With proper hyper-parameter tuning, all seven converge to nearly the same loss at the end of training.
The difference in their per-token losses occurs only in the last few positions. 
We notice that the balanced and aligned topologies typically have a higher loss, while the anti-aligned topologies converge to a lower last-token loss.
However, we perform experiments with $\tau_0$ since it is an interpretable model, as outlined in Section~\ref{sec:interp}.
There is some difference in the training \textit{dynamics}, which is reported in Fig.~\ref{fig:ablation}. The anti-aligned topologies start at a higher loss but converge to a lower loss and vice versa for the algined topologies.

\paragraph{Role of MLPs} MLPs have a significant effect on performance gap to transformers. We compare a $n=32$ \tree with MLPs to a  parameter-matched \tree without MLPs.
For either topology, removing the MLP worsens performance. This effect is more pronounced at higher values of the logit  index. However, the comparison changes both the MLP/nonlinearity and the contraction rank to maintain parameter parity, so it does not isolate which factor causes the gap.
We also observe that MLPs control the variation between different \tree topologies during training. 
We leave an in-depth investigation of the role of MLPs to future work.
However, we would like to emphasize that \textit{even without MLPs, \trees can learn language structure.}

\paragraph{Role of LayerNorm and Residual Connections} LayerNorm and skip connections both help stabilize training. Removing either gave a worse loss (increase of 0.1 to 1.3 nats) for two topologies $\tau_0$ and $\tau_{-1}$, see Table~\ref{tab:ablations}. 
Without either components, the initial gradients are essentially zero (median of order $10^{-17}$ or below, against about $10^{-3}$ for the full model). 
Even the bare \tree (without normalization, residual streams or MLPs) have a median gradient of the order $10^{-12}$ or less.
Therefore, we did not run the latter two combinations at full scale.

\begin{table}[h!]
\centering
\begin{tabular}{lcccc}
\toprule
Topology & Full & No LayerNorm & No residual & No MLP \\
\midrule
aligned $\tau_0$         & 1.847 & 2.567 ($+0.72$)$^{\dagger}$ & 1.946 ($+0.10$) & 1.970 ($+0.12$) \\
anti-aligned $\tau_{-1}$ & 1.805 & 2.025 ($+0.22$)   & 3.142 ($+1.34$) & 1.935 ($+0.13$) \\
\bottomrule
\end{tabular}
\vspace{2pt}
\caption{Best validation loss for $n=32, d=1024$ \tree on TinyStories.  Parameter-matched ablations on various components of \trees. \footnotesize $^{\dagger}$Diverges late in training.}
\label{tab:ablations}
\end{table}

\section{Emergent sparsity through topology routing}
\label{sec:interp}

\begin{figure}[t!]
    \centering
    \includegraphics[width=0.8\linewidth]{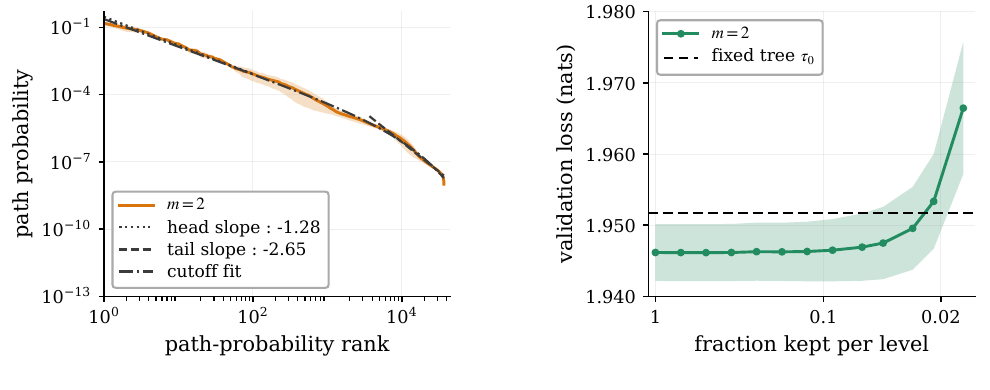}
    \caption{
    Emergent sparsity in \trees. 
    \textbf{Left.} \Trees that are allowed to choose sequences through routing, demonstrate a preference for few topologies. The paths follow a  power-law with a generalized exponential cutoff, with the tail showing a steeper decay.
    \textbf{Right.} Pruning merges at each level can preserve predictive performance even with significant fraction of merges removed. 
    }
    \label{fig:interp}
\end{figure}

To test whether the network develops a preference among topologies, we implement a dynamic \tree with one linear router at each merge layer. The routing decision is sequence-level, unlike token-level choice as in Mixture-of-Experts routing.
At layer $\ell$, the router applies the same linear map to every active node state and then averages across active positions, in the style of \citep{zaheer2017deepsets}:
\begin{equation}
    \label{eq:router}
    r^{(\ell)}_g(x_{1:n-1}) = \frac{1}{|A_\ell|}\sum_{t\in A_\ell}
    w_{\ell,g}^{\top}\operatorname{LN}(z^{(\ell)}_t),\qquad
    p^{(\ell)}=\operatorname{softmax}(\beta r^{(\ell)}+a^{(\ell)}).
\end{equation}
Here $A_\ell$ is the active-node set, $g$ indexes the valid grouping patterns at that layer, and $a^{(\ell)}$ is a non-gradient bias trick adapted from \citep{deepseekai2025deepseekv3technicalreport}.
During training, $a^{(\ell)}$ is updated toward a slightly perturbed uniform selection target to reduce route collapse, which we switch off after 5\% of training time for better performance.
We select the hard top-$m$ indices under $p^{(\ell)}$ and mix their outputs using the corresponding differentiable softmax weights. 
Gradients flow through the selected weights.%
We report the path probabilities $p_{\text{path}} = \mathbb{E}_{x} \left[ \prod_{g_\ell \in  \text{path}} p^{(\ell)}_{g_\ell} \right]$ where $g_\ell$ denotes the topology chosen at a layer $\ell$.

We observe that the network has a sharp cutoff in the probabilities, indicating that it has a strong preference for a few topologies.   
For the specific case of $n=32$, the top-$100$ (out of $36{,}720$ topologies) carry a substantial $O(1)$ fraction of the probability mass.
This can also be seen through the fact that the tail is steeper than the head for both $m$.
Raising $m$ flattens both exponents together, and spreads the mass.
These results are shown in Fig.~\ref{fig:interp}a.
Although the concentration persists despite the balancing bias, current experiments are insufficient to rule out initialization or router-optimization effects as compared to learnt-language preference. 
The exact preferred topologies chosen is an effect of the random seed.
We emphasize that we treat topology as an architectural prior. We do \textit{not} infer linguistic meaning from the studied topologies, as cautioned by \citep{williams2018latent}. 

We test the practical relevance of routing by pruning different fractions of the Merge operations at each level. As one would expect from the concentration of paths, we see that we can remove close to 95\% of the Merge operations, without the \trees showing any meaningful change in performance, cf. Fig.~\ref{fig:interp}b. 
This experiment suggests that we can increase tree-gram capacity by training only sparse subsets of all topologies. 
We leave this to future work.

\section{Conclusion, Limitations, Future Work}

In this work, we present \trees, trainable and controllable $n$-gram approximators of language. They match optimal $n$-gram estimates at small $n$. A noticeable gap from deep models remains at longer contexts.
We find that controlled data approximations can aid hyper-parameter transfer under $\mu$P. We also use them to study performance gaps after finetuning by distinguishing pre-training and fine-tuning distributions.

Since the focus in this work was experimental rather than empirical, our work consequently suffers from many limitations. 
It is not obvious whether our choice of architecture is optimal.
We consider a narrow class of topologies, which is fixed by the layer recurrence relation \ref{eq:nodes} and the maximal order of the $\operatorname{Merge}$ operation. We implement a particular choice of tensor factorization in \ref{eq:merge}, inspired from \cite{loconte2024}.
Computational constraints prevent scaling to longer context windows, particularly to $n \approx 256$, which is the typical length of paragraphs in TinyStories. The $O(Bn^2d)$ activation cost of prefix unfolding is a primary bottleneck, which could be reduced by sharing prefix states.
In addition, the scope and scale of the models and data is limited.
Understanding of different individual components of the \tree architecture also remains limited.
In addition, a careful comparison of alternate HP tuning strategies was not performed.

Developing faster, stabler and robust optimization routines for \trees remains an open engineering problem.
It would also be interesting to train \trees on distributions very different from language.
We leave these and other directions mentioned in the text for future work.

\section*{Declaration of AI usage}

In this work, we used generative AI tools to design or provide feedback on research  methodology or experiments, propose or refine hypotheses, 
help develop theoretical models or conceptual frameworks,  
formulate mathematical claims, 
provide critical ingredients for proving mathematical claims, assist in the writing of proofs,
and to implement several baseline methods. 
We have not used generative AI tools to generate synthetic data sets.
Additionally, we used generative AI tools to 
create or modify scientific figures or images,
create or edit software code,
draft parts of a research paper,
summarize or analyse existing literature,
discover research topics or identify gaps, 
suggest a structure for a research paper, and to 
propose a keywords for a research paper.
All content, claims,
and conclusions have been reviewed and verified by the
authors to ensure accuracy and originality. 
In particular, we checked LLM-generated research ideas for potential plagiarism through a manual literature survey. 
LM-generated code was verified and tested for correctness by one author. 
LM-generated proofs were checked by one author.
The key idea of developing \trees were originated by the authors, and AI was mainly used to implement it from scratch.
We take responsibility for the final content of this work, including text, claims or artifacts produced with the aid of generative AI.

\section*{Ethics Statement}

The authors pledge to adhere to the Code of Ethics. No human subjects were used in this project.
All experiments are conducted exclusively on publicly available, standard benchmark datasets. Specifically, we trained on TinyStories, which is a GPT-generated text dataset WikiText (derived from verified Wikipedia articles) and FineWeb-Edu (an educational subset filtered from Common Crawl). No human subjects were involved, and no private, sensitive, or personally identifiable information (PII) was collected or processed beyond what exists in these public corpora.
\section*{Reproducibility Statement}

Code to train \trees and reproduce the plots in the main text will be  released publicly. Any unlawful or unethical use or distribution of the code is strictly prohibited.

\section*{Acknowledgements}
A.D. acknowledges support from NSF DMR-2429246.
F.C. acknowledges support by the European Union’s Horizon Europe program under the Marie Skłodowska-Curie grant agreement No. 101154584.
D.D. acknowledges financial support from the Johns Hopkins University Data Science and AI Institute Fellowship.
We acknowledge the University of Maryland supercomputing resources (\url{http://hpcc.umd.edu}) made available for conducting the research reported in this paper. 
Certain products, commercial and otherwise, are mentioned in this
publication. 
These mentions are for informational purposes only, and do not imply recommendation or endorsement by NIST.
Contributions to this work by NIST, an agency of the US government, are not subject to US copyright.

\bibliography{iclr2027_conference}
\bibliographystyle{iclr2027_conference}

\appendix

\section{Model, Code and Data release}
\label{app:lic}

The \tree architecture and pre-trained models will be released under the \textbf{Creative Commons Attribution 4.0 International (CC BY 4.0)} license, allowing use, modification and redistribution with proper attribution. 

 The models will be released through a public Github or HuggingFace repository. As presented in the Supplementary Material, scripts describing the architecture and use will be uploaded, with clear instructions for reproducibility.

The Github repo will be additionally tracked to allow users to report and fix bugs. We will further invite the community to contribute to existing models through bug reports, suggestions for improvement and new dataset and feature suggestions. Each update will be accompanied by well-documented release notes, detailing changes and new requirements.

\section{Detailed Background and Related Works}
\label{app:related}

We provide a detailed description of previous and related works, to motivate \trees presented in this work.

\paragraph{Synthetic models of data} 
Modelling language has been an area of study for a long time.
The classic work of \citep{Chomsky1956} used finite-state Markov models, phrase-structure grammars and transformational grammars.
The theory of formal languages classifies grammars by their generative capacity \citep{Chomsky1959}. In particular, context-free grammars use recursive production rules to generate hierarchical and nested structures, such as Dyck languages, that finite-state Markov models cannot represent \citep{ChomskySchutzenberger1963}.
Later work extended context-free grammars into stochastic models by assigning probabilities to production rules \citep{Booth1969,LariYoung1990}, while synthetic grammatical data became a controlled setting for studying whether neural networks can learn hierarchical structure \citep{Elman1991}.
More recently, there has been a growing interest in understanding models of data with randomly chosen underlying rules, eg. the Random Language Model \citep{PhysRevLett.122.128301} and the Random Hierarchy Model \citep{PhysRevX.14.031001}.
The justification for such hierarchical structures comes from theoretical studies \citep{Chomsky1957}, statistical analysis \citep{thurner2015understanding, Goodfellow2016}, information theory \citep{futrell2026linguistic, yang2026functionwordsstatisticalcues} and
cognitive representations \citep{Crain_Nakayama_1987, doi:10.1073/pnas.1018711108, ding2016cortical, hale-etal-2018-finding}. 
\Trees are a natural extension of such hierarchical and exponentially expressive models of language-like data. By making these models trainable, we allow \trees to capture necessary features from the data, without explicit interpretation of hidden nodes.

\paragraph{$n$-gram estimators} 
Language modeling is essentially an $n$-gram estimation task. Recall that the $n$-gram estimation amounts to estimating the conditional probability tensor $p(x_n \mid x_{1:n-1})$.
The frequentist method quickly becomes infeasible due to an exponential dependence on sample complexity and memory requirements.
However, numerous empirical studies indicate that this tensor is extremely sparse, which gives rise to different algorithms. 
Smoothing re-distributes the probability mass through a linear interpolation \citep{chen1996smoothing}; 
PLRE takes advantages of conditional factorization into lower-rank tensors \citep{parikh2014plre}; 
Some algorithms use backoff \citep{goodman2001bitprogresslanguagemodeling, nakamura1989study} 
or scales by brute force \citep{Brants2007LargeLM}
up to the unbounded suffix-array construction of infini-gram \citep{liu2025infinigramscalingunboundedngram}

A different method is to approximate this with factorized tensor sequence models, MPS and tensor-train models. \citep{pestun2017tensor, miller2021tensor, su2024tensortrain}
We use \citep{loconte2025relationshiptensorfactorizationscircuits} as a primary motivation for our rank-$3$ tensor factorization.

Neural networks can also be used for a parametric estimation. This has been tried by \citep{MLPEmulation} with MLPs and one-hot encodings, \citep{bengio2003neural} with MLPs and learnable encodings, \citep{blat2005hybrid} with interpolation with unigrams.

Modern recurrent networks or sequence-to-sequence models take advantage of implicit biases or statistical or nested properties of language to directly model the $n$-gram probabilities. \citep{TfrNGramStats,krishnan2023ngramapproximationpretrainedlanguage,cagnetta2024scaling, cagnetta2025scalinglawsrepresentationlearning}

\paragraph{Controlled approximation of languages}
$n$-gram estimators provide a truncated-horizon approximation of language data. 
In addition, several model families introduce monotonic measures of capacity. 
Weighted automata uses finite Hankel rank \citep{denis2016hankel}.
MPS uses bond dimension \citep{pestun2017tensor,miller2021tensor,su2024tensortrain}
\citep{yang2021pcfgs} uses latent state size, while
\citep{parikh2014plre} use matrix and tensor ranks to represent intermediate levels of dependence between successive $n$-gram orders.

\paragraph{Studying transformers on such controlled languages / n-grams}
There have been attempts to study transformers on synthetic formal grammars.
In addition, $n$-gram models of data has also been formulated
as a learning problem by \citep{svete-etal-2024-transformers} 
and \citep{vare2025learningincontextngramstransformers} in an in-context setting.
\citep{yuksel2026incrementallearningsparseattention} show that transformer incremental learn higher order Markov structure.
However, to the best of our knowledge, such controlled approximations have never been applied for understanding practical situations like HP transfer, due to the limitation posed by scalability and fidelity to a realistic text corpora.

\paragraph{Trainable tree-structured models} 
Differentiable tree structures exist.
In particular, Tree-LSTMs consume a given parse \citep{tai2015treelstm} while \citep{choi2018learning} learns the composition order (tree) with a G\"umbel relaxation.
\citep{shen2019ordered} forces a soft hierarchy into a recurrent net's gates.
A key tension arises with \citep{williams2018latent}, who caution that the structures such models learn need not be the linguistically meaningful ones. 
Transformer architectures also take advantage of trees, eg.
\citep{madaan2023treeformer} uses a decision-tree index over keys.

\section{Details of the architecture}
\label{app:architecture}

\subsection{Analytic expressions for various properties of the network}

Write $n_k$ for the number of nodes at level $k$, so that $n_0 = n$ and
$n_{k+1} = \lfloor n_k/2 \rfloor + 1$. Unrolling the recursion gives the closed form

\begin{equation}
    n_k = 2 + \left\lfloor \frac{n-2}{2^k} \right\rfloor .
    \label{eq:level-size}
\end{equation}

The expressions for the depth $d_n$, number of nodes $s_n$ and the number of different
topologies $t_n$ are as follows

\begin{equation}
d_n=\left\lceil \log_2(n-1)\right\rceil \sim \log_2 n
\label{eq:depth}
\end{equation}

\begin{equation}
s_n= \sum_{k=0}^{d_n} n_k = n + 2 d_n + \sum_{k=1}^{d_n}
\left\lfloor \frac{n-2}{2^k}\right\rfloor
\sim 2n + O(\log n)
\label{eq:num-nodes}
\end{equation}

\begin{equation}
t_n=\prod_{k=0}^{d_n-1}
\binom{n_{k+1}}{\,n_k - n_{k+1}\,}
\sim n^{\log \sqrt{n} + O(1)}
\label{eq:topos}
\end{equation}

Only the $d_n$ levels above the leaves carry weight matrices, so the total
number of parameters is

\begin{equation}
\label{eq:tree-params}
    P_{\text{tree-gram}}(n,r) \approx Vd + d^2 \left( 1 + \sum_{k=1}^{d_n} n_k
    + 2 d_f (2 d_n + 1) \right)
\end{equation}
If MLPs are not included then $d_f = 0$.

\section{Description of $n$-gram baselines used for comparison}
\label{app:baselines}

\paragraph{Raw count}
A \textbf{\textit{raw count}} $n$-gram model estimates $p(x_n\mid x_{1:n-1})$ by normalizing counts for each observed $n$-tuple of tokens.
Smoothed raw count models redistribute probability mass to unseen events and shorter histories through interpolation and absolute discounting, while low-rank ensembles control the correlation of $n$-grams through low-rank matrices \citep{chen1996smoothing,parikh2014powerlowrank}.
We report classical raw-count, smoothed-count, and low-rank count baselines only for $n=2$ to $n=8$. 
This is because at $n=8$, we already observe that 60\% of the $n$-grams appear only once.
\Trees match these tested count baselines at small $n$ and out-perform them for larger $n$.

\paragraph{Infini-gram}
Infini-gram uses a suffix array to retrieve continuation counts for the longest matching suffix of the available history and backs off when longer matches are unavailable \citep{liu2025infinigramscalingunboundedngram}.
In our finite-context comparison, we cap this search at $n-1$ tokens and interpolate with a smoothed unigram  (\textbf{\textit{infinigram + unigram}})
or a neural LM (\textbf{\textit{infinigram + neural LM}})
\unskip. On TinyStories, infinigram + unigram improves through $n=4$, but its loss worsens and then plateaus at larger $n$ as long exact suffix matches become rare.
Infinigram + neural LM improves but plateaus by $n \approx 8$.
In contrast, \trees improve throughout the tested range and outperform infinigram at every shared value of $n$.

\paragraph{Spectral automaton models}
A spectral weighted-finite-automaton (\textbf{\textit{spectral WFA}}) estimator constructs empirical Hankel blocks of substring probabilities and uses a truncated singular-value decomposition to recover a finite-dimensional state and token-transition operators \citep{balle2014spectral}.
After tuning the rank and Hankel window $W\in\{3,\ldots,6\}$, the resulting curve remains nearly flat as $n$ increases and performs worse even compared to the raw trigram.
\Trees outperform the spectral estimator at every shared order, with the gap increasing toward larger $n$.

\paragraph{Tensor train approximations}
We compare to three tensor-network language models. The uniform matrix product state (\textbf{\textit{u-MPS}}) of \citep{miller2021tensor} is a Born machine, $p(x_n\mid x_{1:n-1})\propto|\alpha^\top A^{x_1}\cdots A^{x_n}\omega|^2$, with a single core $A\in\mathbb{R}^{|V|\times D\times D}$ shared across positions. The tree tensor network (\textbf{\textit{tree TN}}) of \citep{cheng2019tree} is a Born machine on a balanced binary tree over the context, with a learned $|V|\to v$ feature map at the leaves and one order-3 tensor at each internal node; we train it end to end by gradient descent rather than by DMRG sweeps. TTLM-Large (\textbf{\textit{TTLM}}) \citep{su2024tensortrain} is a tensor-train recurrence in which a token-dependent factor updates the state multiplicatively, followed by a softmax readout tied to the input embedding tensor.
At every $n$, the bond dimension $D$, the tree width $v$ and the rank $R$ are set to match the parameter count of the $d=1024$ \tree ($D=47$--$131$, $v=164$--$316$, $R=46$--$128$). The tree TN is best at $n=3$ ($3.36$ nats) and falls to near the unigram level from $n=8$ on. \Trees outperform all three at every shared order, by $0.19$ nats at $n=3$ and $0.95$ nats at $n=32$ against the strongest of them, TTLM.

\paragraph{Neural network approximation of $n$-grams}
A neural $n$-gram model embeds a fixed window of $n-1$ preceding tokens and uses an MLP with a softmax output to estimate the next-token distribution. This may be done with one-hot encodings \citep{MLPEmulation} (\textbf{\textit{MLP}}) or through learnable embedding structure \citep{bengio2003neural} (\textbf{\textit{neural LM}}).
Later works combined MLP with a unigram model \citep{blat2005hybrid} which we adapt here as the \textbf{\textit{MLP + unigram}} model with the learnable embedding MLP.
The parameter-matched neural baseline improves through $n=16$ but worsens at $n=32$. However, it has consistently higher cross-entropy than \trees.
Thus, \trees outperform all tested finite-window neural $n$-gram estimators at every shared order.

\paragraph{Deep neural sequence models}
Well-trained deep sequence models provide a stronger estimate of $n$-gram statistics at large $n$ \citep{TfrNGramStats,krishnan2023ngramapproximationpretrainedlanguage,cagnetta2024scaling}.
In this case, we compare \trees to GPT-2 with absolute positional embeddings (\textbf{\textit{GPT-2 (APE)}}) and rotary positional embeddings (\textbf{\textit{GPT-2 (RoPE)}}), Llama-3.2 (\textbf{\textit{Llama}}), and Mamba (\textbf{\textit{Mamba}}).
These are converged multi-epoch runs, which use approximately six times the token budget.
In Fig.~\ref{fig:scaling}b, \trees track these estimators through approximately $n=8$, while the deep references remain lower at larger $n$.
However, in Fig.~\ref{fig:intro}(c), cross-entropy decreases monotonically as the rank $d$ increases over the tested range.

\section{A Toy Model for \Tree data}
\label{app:transfer-toy-model}

\input{appendix_dyadic_toy_model}

\section{\Tree experimental details and comparison to transformers}
\label{app:exp}

We performed experiments with 4 NVIDIA A100 GPUs. The 529.2M-token run for the $n=32$, $d=1024$ model (with MLP) takes 5 hours of wall-clock time. A total of 11,000 A100 GPU hours were used for all pre-training experiments on \trees reported in this paper.

We train on the $529.2$M-token training split of TinyStories, tokenized with a custom BPE tokenizer \citep{cagnetta2024scaling}, and validate on a $1.05$M-token subset of its $5.34$M-token validation split.

Every run trains for a single epoch over the $529.2$M training tokens, cut into non-overlapping windows of $n$ tokens with $1024$ windows per batch, using AdamW at its default moments. The learning rate warms up linearly from the initial to the peak value over the first $1\%$ of steps and then follows a cosine decay to $1\%$ of the peak. Unless stated otherwise the architecture uses LayerNorm, a residual stream of unit strength, GELU, $d_{\mathrm{ff}} = 4d$ and tied embeddings. We used a 1.05M validation set.

We tuned the peak learning rate and the weight decay for every cell. The $n$-sweep used $\eta_{\max} \in \{3\!\times\!10^{-5}, 10^{-4}, 3\!\times\!10^{-4}, 10^{-3}, 3\!\times\!10^{-3}\}$ and $\lambda \in \{10^{-3}, 10^{-2}, 10^{-1}, 1\}$, and the $d$-sweep added $\eta_{\max} = 10^{-2}$; the $n=32$, $d=1024$ cell pools this grid with an earlier one that reaches $\eta_{\max} = 3\!\times\!10^{-2}$ and $\lambda = 10^{-4}$. The initial learning rate was $10^{-5}$ except for $\tau_{-1}$ and $\tau_{-2}$. The topology ablations used smaller grids to test for learning rate transfer: $\eta_{\max} \in \{3\!\times\!10^{-4}, 10^{-3}, 3\!\times\!10^{-3}\}$, with $\lambda \in \{10^{-3}, \ldots, 1\}$ for $\tau_1, \tau_2$ and $\lambda \in \{10^{-2}, 10^{-1}, 1\}$ for $\tau_a, \tau_b$. The anti-aligned $\tau_{-1}$ and $\tau_{-2}$ need smaller rates and were swept over initial rates $\{10^{-6}, 10^{-5}\}$, $\eta_{\max} \in \{10^{-5}, \ldots, 10^{-3}\}$ and $\lambda \in \{10^{-4}, \ldots, 10^{-1}\}$ ($\tau_{-2}$: $\eta_{\max} \in \{10^{-4}, 3\!\times\!10^{-4}, 10^{-3}\}$, $\lambda \in \{10^{-2}, 10^{-1}\}$). Without MLPs only $\tau_0$ and $\tau_{-1}$ were swept; the other five topologies were run at a single setting ($\eta_{\max} = 10^{-3}$, $\lambda = 0.1$). Cells were scored by the minimum validation loss over the run, averaged over two seeds. At the selected settings of the cells with MLPs, the seed-to-seed standard deviation is at most $0.0025$ nats ($d=64$), and at most $0.0004$ nats for the other $n$- and $d$-sweep cells; it reaches $0.06$ nats for the untuned topologies without MLPs.  We ran 2 random seeds for each HP configuration of the $n$- and $d$-sweeps and then ran a total of 4 random seeds for the model with the best HP configuration. Data generation took 200 GPU hours.

For $\mu$P transformers comparison, we scanned 10 peak learning rates while holding the weight decay fixed at $0.01$, and kept the rate with the lowest validation loss on a single seed. We ran a total of 3 seeds for the transformers with the best value of the hyper-parameter combination. These were also trained with AdamW at batch $1024$ for one epoch on the same windows as the \trees, with the same $1\%$ warmup from $10^{-5}$, decaying to $10\%$ of the peak, and with dropout $0.0$. The transformers were based on \citep{kalra2025nanogptmup}.

\Trees were trained on WikiText-103 (130 MT) and FineWeb-Edu (1.05BT subset) as well. The training was done by choosing the best hyper-parameter for the relevant models for the TinyStories corpus, without additional hyper-parameter tuning.
Training time was approximately linear in total token count, except for the $n=32, d=1024$ \tree model which could not be fit into 4 A100 GPUs. As a result we used gradient accumulation, which made the total wall-clock time to be closer to 20 hours, rather than the expected 10 hours.
The additional experiments reported for these datasets needed an additional 8,000 GPU hours. At the time of writing, the pre-training, fine-tuning results on FineWeb-Edu are yet to run, and they need an estimated 2,000 GPU hours more.

\paragraph{Cost of generating a corpus.}
We timed the generation costs for the largest generators in the paper: 139.5 million parameters over an 8{,}192-token vocabulary (TinyStories and WikiText) and 182.6 million parameters over a 50{,}257-token vocabulary (FineWeb-Edu).
On four threads of a single CPU (an Apple M4 Pro), sampling 512 sequences in parallel, they produce about 1.15 thousand and 0.65 thousand tokens per second.
Generating WikiText (131.8 million tokens), TinyStories (529.2 million) and FineWeb-Edu (1.05 billion) therefore takes about 32, 128 and 448 hours on \textit{one} worker.
The deepest KV-cached transformers typically take around 2x less time for generation.

We also measured the estimated generation times on a cluster with dual AMD 7763 CPUs. The times are mentioned below in Table~\ref{tab:gen-cost}. As can be observed, the maximum time for generation even for the 1.05BT subset is roughly 90 hours. The mean times are well within a day for Wiki-Text and TinyStories, and is equal to one and a half days for FineWeb-Edu (which has a larger vocabulary size).

\begin{table}[t]
\centering
\small
\setlength{\tabcolsep}{4pt}
\resizebox{\textwidth}{!}{%
\begin{tabular}{cl rrr rrr rrr}
\toprule
 & & \multicolumn{3}{c}{WT (130\,MT)} & \multicolumn{3}{c}{TS (529\,MT)} & \multicolumn{3}{c}{FWE (1.05\,BT)} \\
\cmidrule(lr){3-5}\cmidrule(lr){6-8}\cmidrule(lr){9-11}
Cores & Model & median & mean & max & median & mean & max & median & mean & max \\
\midrule
\multirow{2}{*}{1}  & \Tree        & 58.3 & 91.5 & 320  & 232  & 362  & 1{,}253 & 956  & 1{,}097 & 2{,}985 \\
                    & Transformer & 40.4 & 50.4 & 125  & 163  & 203  & 499     & 779  & 778     & 1{,}543 \\
\midrule
\multirow{2}{*}{16} & \Tree        & 3.9  & 6.2  & 21.2 & 15.7 & 24.3 & 83.6    & 70.2 & 80.2    & 205 \\
                    & Transformer & 2.8  & 3.4  & 8.3  & 12.0 & 14.6 & 35.1    & 58.2 & 58.0    & 104 \\
\midrule
\multirow{2}{*}{32} & \Tree        & 2.3  & 3.7  & 12.3 & 9.4  & 14.0 & 46.4    & 41.2 & 48.9    & 124 \\
                    & Transformer & 1.7  & 2.0  & 4.6  & 6.6  & 7.7  & 18.0    & 33.2 & 33.3    & 58.4 \\
\midrule
\multirow{2}{*}{64} & \Tree        & 1.7  & 2.9  & 9.2  & 6.8  & 10.3 & 32.5    & 31.3 & 36.2    & 88.7 \\
                    & Transformer & 1.4  & 1.5  & 3.3  & 4.2  & 4.9  & 11.1    & 24.3 & 24.6    & 40.0 \\
\bottomrule
\end{tabular}}
\caption{Wall-clock time for corpus generation on dual AMD 7763 CPU cores. Times are in units of hours. Transformers are typically around 1.5-2x faster.}
\label{tab:gen-cost}
\end{table}

\subsection{Parameter-matched transformer configurations}
\label{app:matched-configs}

Table~\ref{tab:matched-configs} gives the transformer width $D$ used for each \tree and
transformer depth in Fig.~\ref{fig:treevstfr}. All transformers use eight heads, MLP width
$4D$, GELU, RoPE, tied embeddings, and the same context and token budget as the corresponding
\tree. Parameter counts are matched to the nearest available width. Thus, width varies with
depth and the sweep does not isolate depth.

\begin{table}[h]
\centering
\caption{Exact parameter-matched transformer widths. Tree parameters include embeddings.}
\label{tab:matched-configs}
\scriptsize
\resizebox{\textwidth}{!}{%
\begin{tabular}{rrrrrrrrrrr}
\toprule
\multicolumn{2}{c}{\tree} & Tree params & \multicolumn{8}{c}{Transformer width $D$ by depth $L$} \\
$n$ & $d$ & & $1$ & $2$ & $3$ & $4$ & $5$ & $6$ & $10$ & $12$ \\
\midrule
3  & 1024 & 36,709,376  & 1440 & 1072 & 896  & 800  & 720  & 656  & 512  & 480 \\
4  & 1024 & 56,637,440  & 1856 & 1376 & 1152 & 1008 & 912  & 832  & 656  & 592 \\
8  & 1024 & 78,665,728  & 2240 & 1648 & 1376 & 1200 & 1072 & 992  & 784  & 704 \\
16 & 1024 & 104,892,416 & 2640 & 1920 & 1600 & 1392 & 1248 & 1152 & 896  & 832 \\
32 & 64   & 1,039,744   & 112  & 96   & 96   & 80   & 80   & 80   & 64   & 64 \\
32 & 128  & 3,103,488   & 272  & 224  & 208  & 176  & 176  & 160  & 128  & 128 \\
32 & 256  & 10,302,976  & 640  & 512  & 432  & 384  & 352  & 320  & 256  & 240 \\
32 & 512  & 36,989,952  & 1440 & 1088 & 912  & 800  & 720  & 656  & 528  & 480 \\
32 & 1024 & 139,515,904 & 3088 & 2240 & 1856 & 1616 & 1456 & 1344 & 1040 & 960 \\
\bottomrule
\end{tabular}}
\end{table}

\section{HP transfer experiment details}
\label{app:hp}

\subsection{Learning rate sweeps}
\label{app:hp-sweeps}

For the hyper-parameter-transfer study in Fig.~\ref{fig:hpt}, all transformers use eight heads, $d_{\mathrm{ff}}=4d$, GELU, RoPE, tied embeddings, no dropout, sequence length $128$ and batch $256$ ($32{,}768$ tokens per step). Every proxy and every target trains for one epoch over the full $529.2$M-token training split of its corpus ($16{,}024$ steps), so a proxy trained on real data differs from its target only in width. We use AdamW with $(\beta_1,\beta_2)=(0.9,0.95)$, $\epsilon=10^{-8}$ and weight decay $0.01$, a linear warmup from $10^{-5}$ over the first $1\%$ of steps, and a cosine decay to $10\%$ of the peak.

Proxy widths are $d\in\{64,96,128,160,192,256\}$ at $L\in\{2,4\}$, each trained on all eleven corpora of Fig.~\ref{fig:ptft}. Targets are trained on real TinyStories only, at $d\in\{384,512,640,832,1024,1280,1600,2048\}$ for $L=2$ and $d\in\{288,384,480,608,768,928,1152,1408\}$ for $L=4$; the main text reports $d=1024$ ($L=2$) and $d=1408$ ($L=4$). Peak rates lie on the lattice $\eta_k=0.0039\cdot10^{k/8}$. For the proxies, seed $0$ spans $[0.0039,3.9]$ at $d=64$ and $[0.00925,1.64]$ at the wider widths, and seeds $1$ and $2$ span $[0.052,1.64]$. 
This was done to ensure that the optimal learning rate is obtained at the basin of the loss vs learning rate curve.
Targets use three seeds on $[0.052,6.94]$, or on $[0.0123,2.19]$ for the two widest targets at each depth, plus two to four off-lattice rates per target inserted near its divergence onset.

\subsection{Continuous $\eta^\star$ from a smoothing spline}
\label{app:hp-spline}

In order to get a more precise estimate of $\eta^\star$ we use a cubic smoothing spline as discussed in \citep{kalra2026quantifyinghyperparametertransferimportance}.
Writing $\nu=\log\eta$, we discard points with $L>1.35\,L^\star$, fit a cubic smoothing
spline to $L(\nu)$ with smoothing parameter $s=0.1\,N\,\mathrm{Var}(L)$, evaluate it
at $400$ uniformly spaced $\nu$ in the retained range, and take its minimizer.

Two adaptations are needed here. First, the $1.35\,L^\star$ window was calibrated on sweeps with a different loss scale; with $L^\star\approx1.48$ nats the bar sits near $2.0$ and admits points already far up the divergence cliff. A single such point inflates $\mathrm{Var}(L)$, hence $s$, and the cubic cannot turn fast enough to follow the cliff, so it smears it back across the basin. We therefore truncate the window at the divergence onset, cutting at the first increment above $\max(3h,\,0.05)$ nats, where $h$ is the median $|\Delta L|$ along the descending branch; across all $412$ sweeps no descending increment exceeds $0.045$ nats while the smallest increment the rule flags is $0.062$, so it separates basin from cliff without ever truncating a basin, and the median residual falls to $0.0009$ nats. Second, we use the spline only to \textit{locate} optima. A proxy may recommend an $\eta$ far up the target's cliff --- tuning on the real corpus does exactly that --- and there the measured loss is precisely the quantity of interest, so we score the target with a shape-preserving (PCHIP) interpolant of its full curve rather than with the spline. That interpolant cannot itself locate an optimum, since it sets a zero derivative at every data extremum and its minimiser is the grid argmin again; only a smoothing
spline moves $\eta^\star$ off the lattice.

The re-estimate leaves our conclusions unchanged and sharpens two of them. The median-over-synthetics rule stays within one grid point of each target's own optimum with no grid involved: $\eta^\star_{\mathrm{median}}/\eta^\star_{\mathrm{target}}$
is $0.87$--$1.05$ at $L{=}2$ and $1.11$--$1.22$ at $L{=}4$, against a rung of $4/3$.

\section{Finetuning experiment details}
\label{app:ptft}

\subsection{Methodology and optimization}

For the finetuning experiment the downstream models are $(L,d)=(2,1024)$ and $(4,768)$. We evaluate every ordered pair among the bigram corpus, the five horizon-sweep corpora $\{n3,n4,n8,n16,n32\}_{d=1024}$, the five capacity-sweep corpora $n32_{d\in\{64,128,256,512,1024\}}$, and real TinyStories: an $11\times11$ matrix for each depth and three seeds. Fine-tuned models are validated on a $1.05$M-token subset of the target's validation split.

\textit{Pre-training.} Each source is pre-trained for one epoch over its $80\%$ segment ($12{,}819$ steps). Its peak rate $\eta^\star_{\mathrm{PT}}$ is the argmin, on one seed, over seven lattice rates centred on that corpus's optimum at $d=256$ in the HP-transfer sweep, a $5.6\times$ window; no selected rate lies on the edge of its window. This gives $11\times2\times3=66$ pre-trained checkpoints.

\textit{Fine-tuning.}
All parameters are updated from the pre-trained weights with a fresh AdamW state; the learning rate is re-warmed from $10^{-5}$ over the first 1\% of steps and cosine-decayed to 10\% of its peak \cite{gupta2023rewarm, ibrahim2024simple}, with no replay of source data. The fine-tuning peak rate is set per (source, target) pair: $\eta_{FT}(s \to t) = \eta^\star_{PT(t)} \cdot \eta^\star_{FT(s \to \text{real})}/\eta^\star_{PT(\text{real})}$, snapped to the lattice, where $\eta^\star_{FT(s \to \text{real})}$ comes from a one-seed sweep over a shared 22-point window. Tuning the rate per model follows \cite{isik2025scaling} ; re-warming to the pre-training peak instead trades adaptation against forgetting \cite{gupta2023rewarm}, and we score only the target.

\textit{Budgets and references.} The matrix is read after the full $105.8$M fine-tuning tokens ($3{,}205$ steps). For a synthetic target, the floor is its generator's entropy rate $H_q$ (Eq.~\ref{eq:bayes-floor}). For real TinyStories we use $1.242$ nats, the converged loss of the 12-layer, $d=1024$ GPT-2 of \citep{cagnetta2024scaling} at $T=128$, which is an empirical reference rather than a Bayes floor. Each cell reports the median over the three seeds of the final validation loss. 
As a no-pre-training reference, the same models trained from scratch on the same $105.8$M real tokens, at $0.52$ (within one lattice step of their own optimum on this budget), reach $1.688$ nats at $L=2$ and $1.600$ nats at $L=4$ (medians of three seeds), that is $0.36$ and $0.29$ on the scale of Fig.~\ref{fig:ptft}(c,d).

\section{Additional plots}
\label{app:extra}
\label{app:treevstfr}
\label{app:transfer}
\label{app:dynamics}
\label{app:emb}
\label{app:spectra}
\label{app:extraptft}

\begin{figure}[h]
\centering
\includegraphics[width=0.8\textwidth]{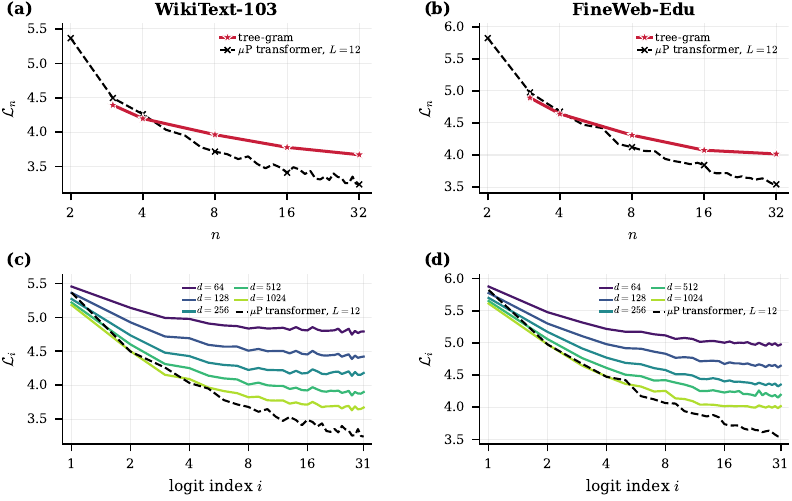}
\caption{Comparison of \trees with transformers on Wiki-Text and FineWeb-Edu. We notice that \trees are competitive with these deep neural architectures, like TinyStories discussed in the main text.
We use a deep well-trained transformer as a proxy for the true empirical $n$-gram entropy.
}
\label{fig:xtra-corpus}
\end{figure}

\begin{figure}[h]
\centering
\includegraphics[width=\textwidth]{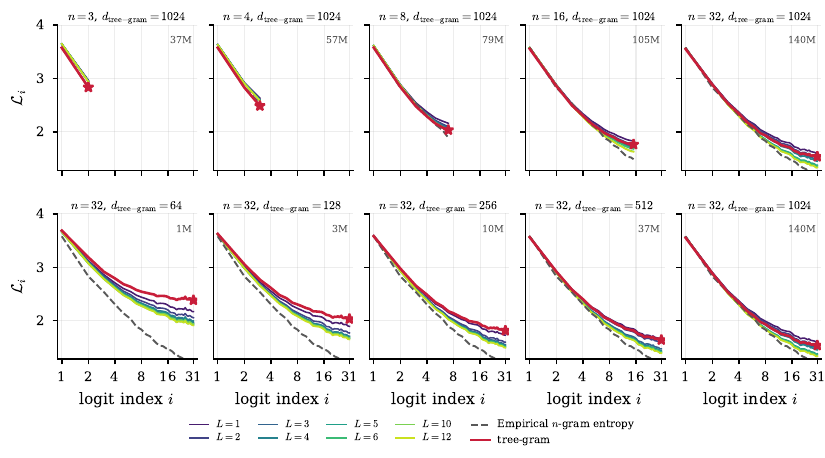}
\caption{Per-position cross-entropy for every \tree in the sweep against its parameter-matched transformers, for various depths $L$. 
\textbf{(top)} $n$ varies at $d=1024$; \textbf{(bottom)} $d$ varies at $n=32$. 
Grey annotations give the parameter count of the \tree in that panel ($1$M to $140$M).
The \tree tracks the transformer family exactly for the first few positions in every panel and peels away only at long context.
}
\label{fig:treevstfrgrid}
\end{figure}

\begin{figure}[h]
\centering
\includegraphics[width=\textwidth]{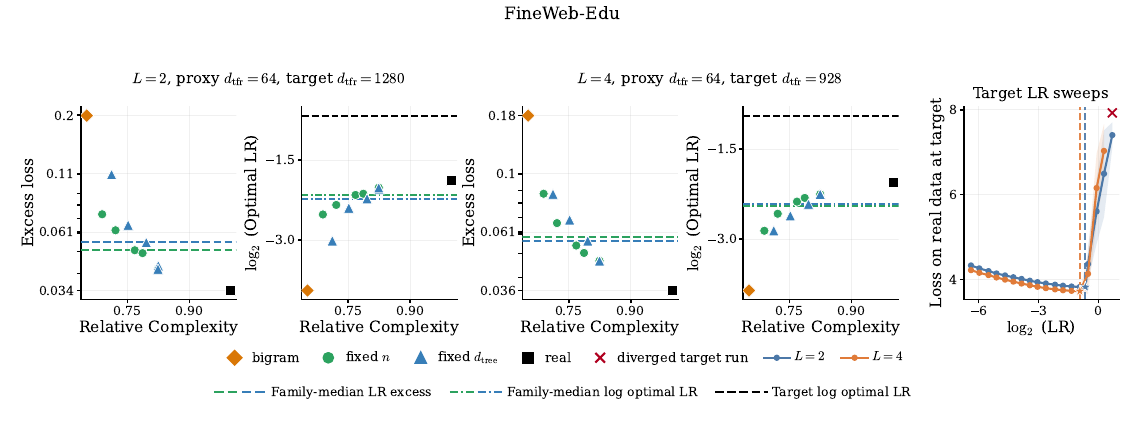}
\caption{
The HP transfer plot for FineWeb-Edu. We notice that in this case, the synthetic corpora do \textit{not} provide a better approximation of the target.
}
\label{fig:hpt-fwe}
\end{figure}

\begin{figure}[h]
\centering
\includegraphics[width=\textwidth]{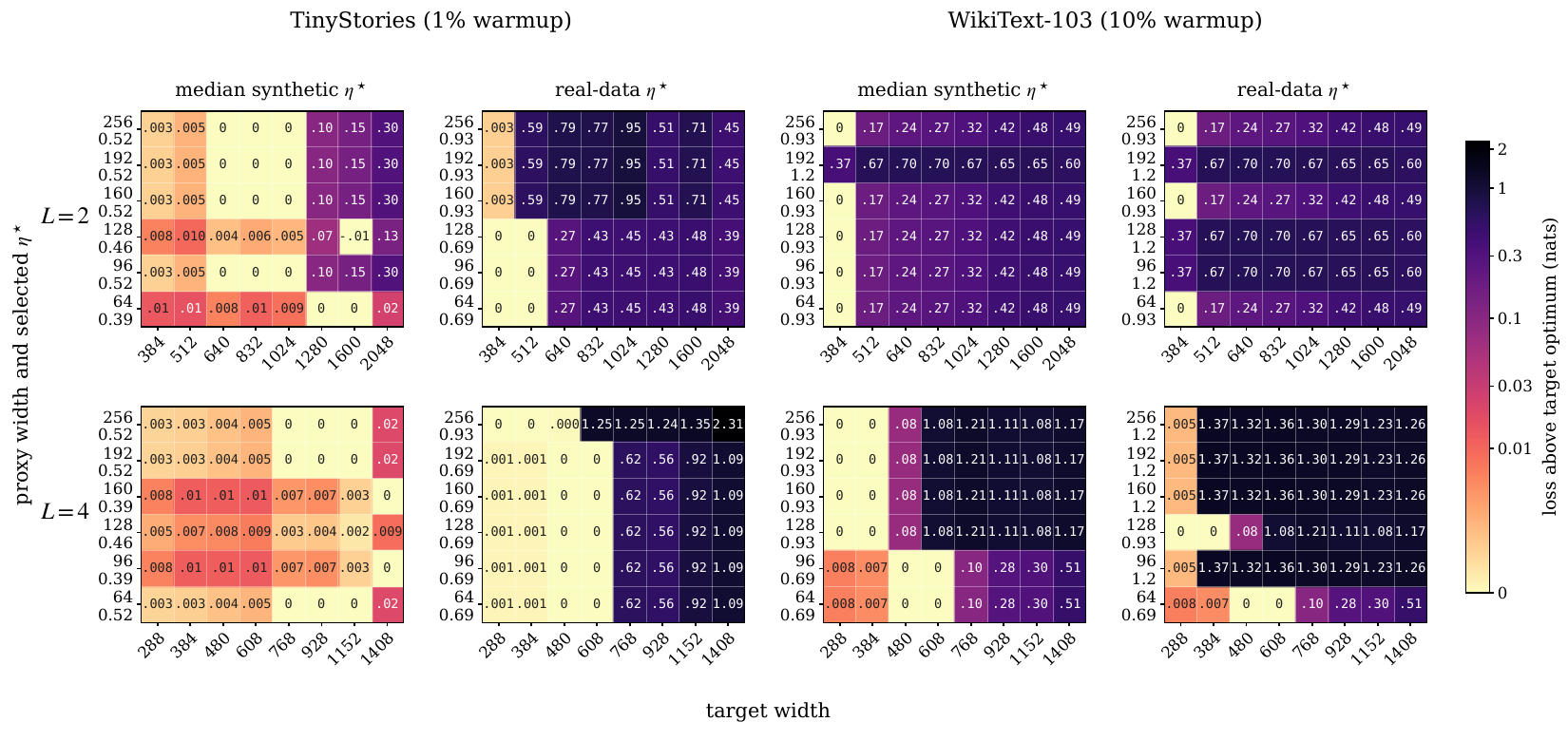}
\caption{
The HP transfer phase diagram. 
(\textbf{Left}) We plot the excess loss obtained when transferring the median optimal learning rate $\nu^\star$ from a source width $d_{\text{source,tfr}}$ to a target width $d_{\text{target,tfr}}$.
(\textbf{Right}) The same plot when we tune a proxy on the real data.
Rows are depth $L=2$ (top) and $L=4$ (bottom). 
The synthetic rule is more robust as compared to tuning on the real data (the standard recipe).
}
\label{fig:transfergrid}
\end{figure}

\begin{figure}[h]
\centering
\includegraphics[width=\textwidth]{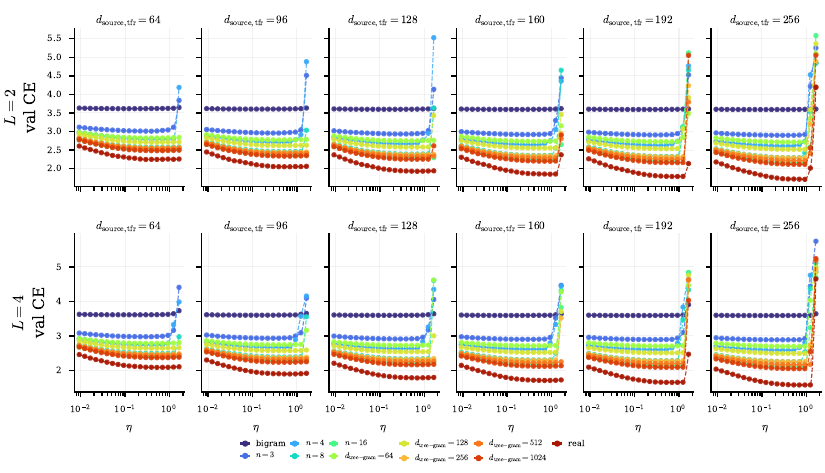}
\caption{Converged losses as a function of the peak learning rates. The solid dots represent the actual runs while the dashed lines represent the interpolated losses using a spline function.
}
\label{fig:losslr}
\end{figure}

\begin{figure}[h]
\centering
\includegraphics[width=\textwidth]{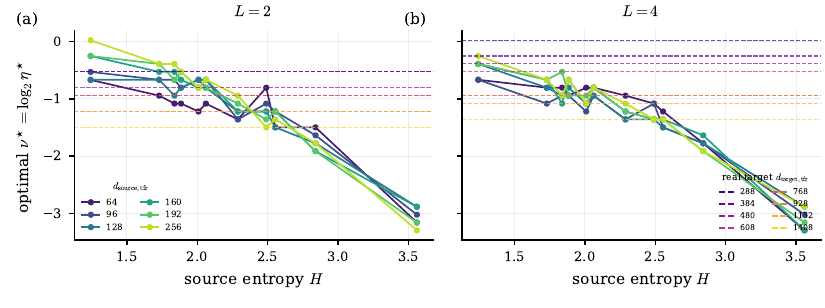}
\caption{
Optimal learning rate $\nu^\star \equiv \log_2 \eta^\star$ as a function of the entropy of the generated distributions. We see a monotonic decrease with increase in entropy. However the width-dependence is extremely weak. 
}
\label{fig:opt-entropy-synth}
\end{figure}

\begin{figure}[h]
\centering
\includegraphics[width=\textwidth]{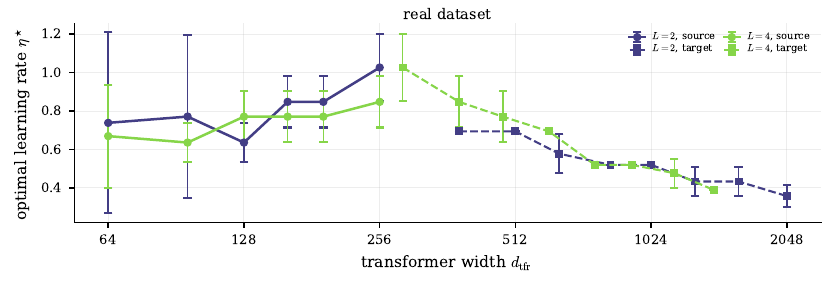}
\caption{
Optimal learning rate $\eta^\star$ for the real dataset as a function of width. We observe a non-monotonic trend.
}
\label{fig:opt-real}
\end{figure}

\begin{figure}[h]
\centering
\includegraphics[width=\textwidth]{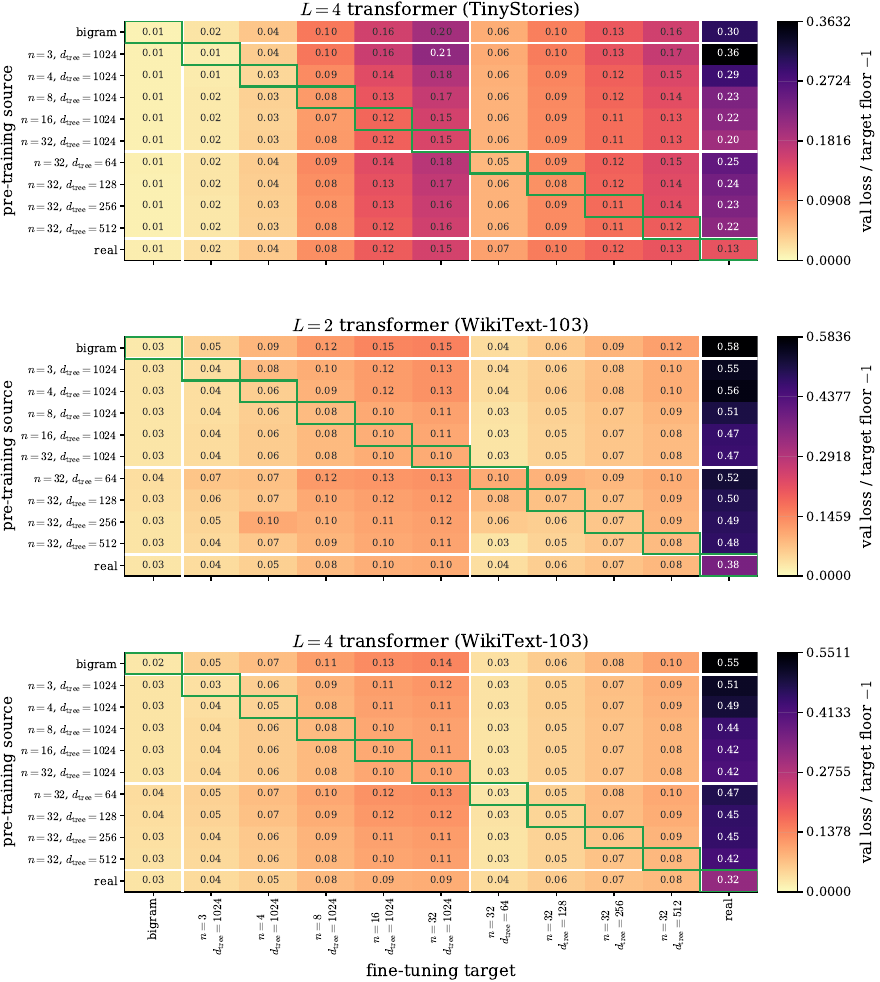}
\caption{
The pre-training transfer matrix for $L=2$ transformer on WikiText and $L=4$ transformers on WikiText and TinyStories respectively. We see the same qualitative trend as that of the main-text figure.
}
\label{fig:ptftextra}
\end{figure}

\begin{figure}[h]
\centering
\includegraphics[width=\textwidth]{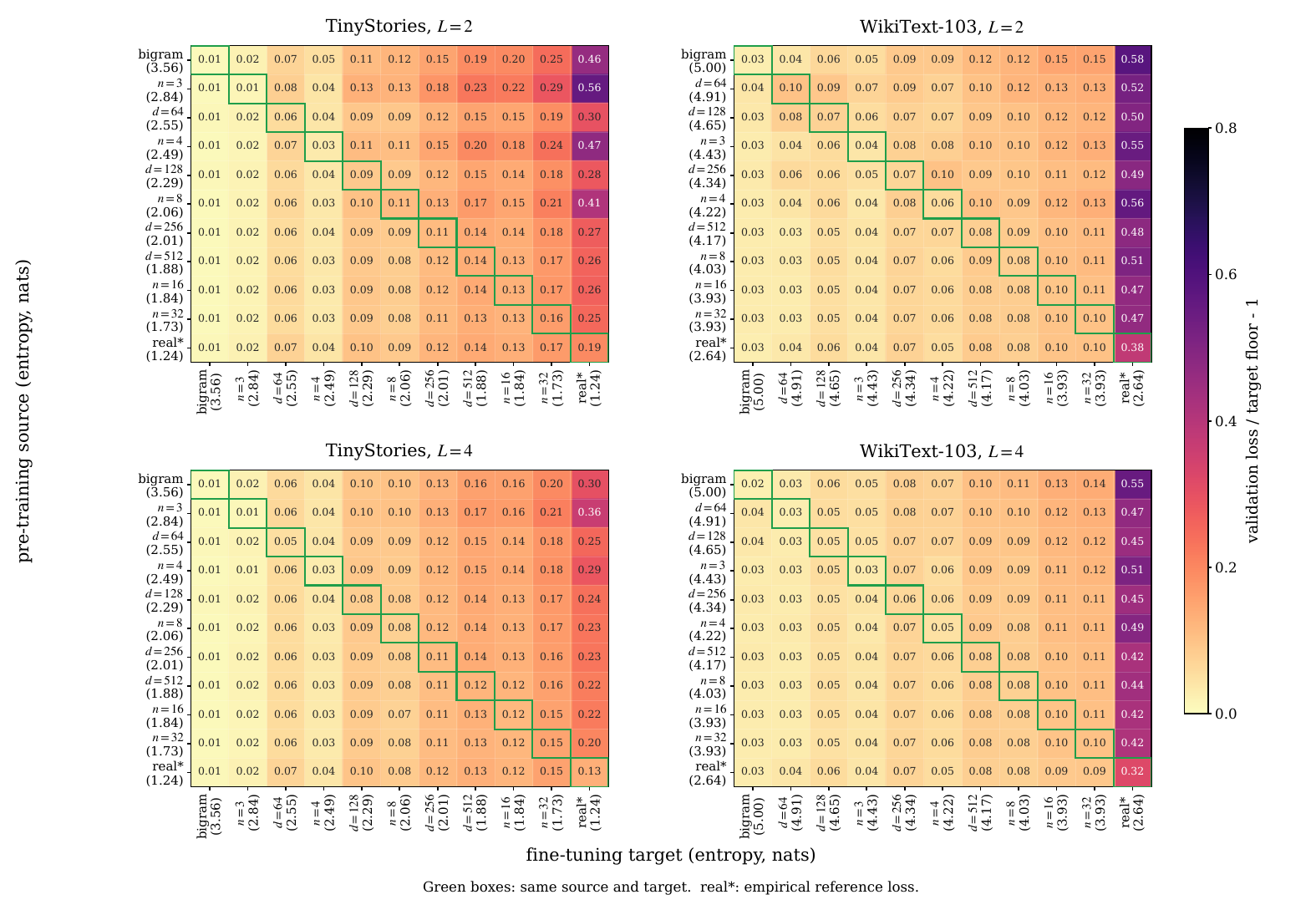}
\caption{
The pre-training transfer matrices with
both axes re-sorted by \textit{entropy rate} of the synthetic corpus, instead of grouped by mode of approximation.
The green boxes are the same-distribution diagonal.
We observe that the monotonic trend we observed in the main text, is no longer present. In addition, horizon-limited modes perform worse than capacity-limited modes, even when they have similar entropy rate.
}
\label{fig:ptftentropy}
\end{figure}

We provide additional plots alluded to in the main text. In order:

\begin{itemize}
\item Comparison of \trees ran on Wiki-Text and FineWebEdu, with deep neural baselines (Fig.~\ref{fig:xtra-corpus}).
\item A per-position cross-entropy comparison of every \tree in the sweep against itsparameter-matched transformers, at every depth (Fig.~\ref{fig:treevstfrgrid}).
\item The HP tuning results on FineWeb-Edu (Fig.~\ref{fig:hpt-fwe}).
\item The transfer phase diagram over every (proxy width, target width) pair, contrasting the median-over-synthetics rule with tuning on the real corpus (Fig.~\ref{fig:transfergrid}). %
\item The raw loss-versus-learning-rate curves for different widths on different datasets, with the spline-function interpolated values. (Fig.~\ref{fig:losslr}).
\item Plots of the optimal learning rate for proxy widths  as a function of the entropy of the distribution is presented in Figure~\ref{fig:opt-entropy-synth}. The non-monotonic behavior of the optimal learning rate as a function of the width for the real dataset is presented in Figure~\ref{fig:opt-real}.
\item The pre-training and fine-tuning transfer matrices for both TinyStories and Wiki-Text dataset (Fig~\ref{fig:ptftextra}).
\item The pre-training transfer matrix with both axes re-sorted by entropy rates, which is where the two synthetic sweeps are seen to interleave (Fig.~\ref{fig:ptftentropy}).
\end{itemize}

\clearpage
\newpage

\section{Pre-training Fine-tuning Performance Gap}
\label{app:ptft-law}
\begin{figure}
    \centering
    \includegraphics[width=\linewidth]{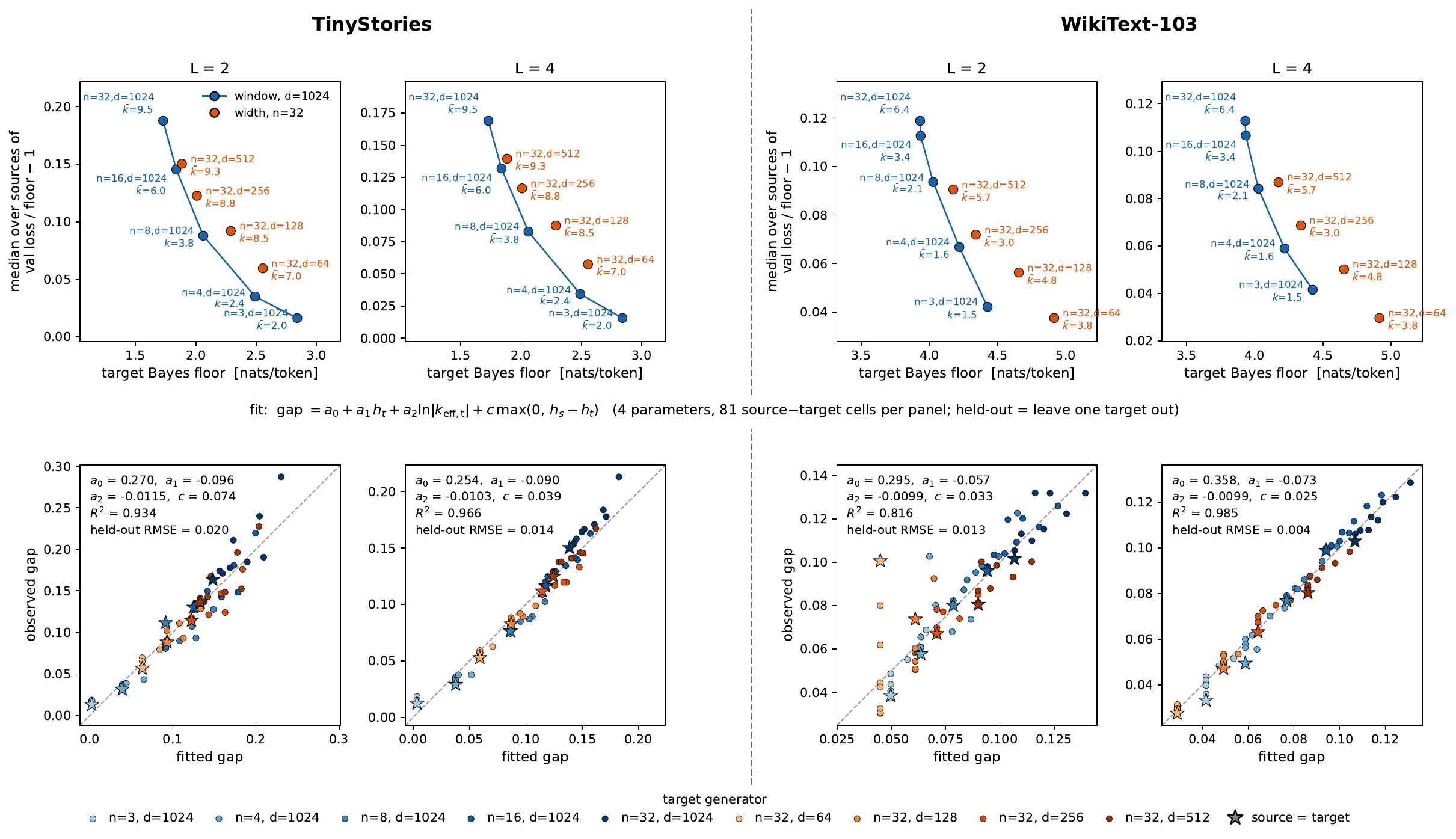}
    \caption{
    Do targets with longer-range correlations have a larger fine-tuning gap at the same entropy? The gap is the final validation loss divided by the target's Bayes floor, minus one. \emph{Top:} median gap over sources against the target's Bayes floor, for generators that vary the context window at $d=1024$ (blue) and generators that vary the width at $n=32$ (orange); numbers give each target's mean lag $\bar k$. If range matters beyond entropy, the orange curve lies \emph{above} the blue one at matched entropy, which it does in every panel. \emph{Bottom:} observed against fitted gap for all 81 source--target pairs under the law shown, coloured by target; stars mark source $=$ target, where the penalty term vanishes. Held-out error is computed by leaving out each target in turn. Points on the diagonal mean the law captures the pair. The one clear miss (WikiText-103, $L=2$, the narrowest-width target trained on itself) is a run that became unstable late in training.
    }
    \label{fig:ptft-law}
\end{figure}

Which source--target pairs are hard to fine-tune is decided almost entirely by the \emph{target}, through two properties measured directly on its token statistics. The first is its entropy rate: targets with a lower Bayes floor are harder to reach relative to that floor. The second is its \emph{effective range}, which is how far apart two tokens can be and still carry information about each other. Entropy has the larger effect, but \emph{at comparable entropy} a target with a longer effective range is consistently harder to reach. Generators that change the width at a fixed window sit above generators that change the window, at the same entropy (Fig.~\ref{fig:ptft-law}, top). The source matters mainly when it carries \emph{less} information than the target. A four-parameter law in these quantities, linear in the target's entropy and in the logarithm of its range, plus a penalty for a source whose entropy rate exceeds the target's, accounts for 82--99\% of the variance across all 81 source--target pairs in each setting. It predicts the gap for held-out targets with comparable error (Fig.~\ref{fig:ptft-law}, bottom). The range term has the same sign and nearly the same size in all four settings.

To measure the effective range, we estimate the mutual information between two tokens $k$ positions apart in text sampled from each generator. We use a bias-corrected plug-in estimator and subtract a null computed on shuffled pairs. Beyond the generator's context window ($k > n$), the logarithm of this mutual information falls linearly with $k$, so the information decays exponentially. We take the slope of that line, $k_{\rm eff}$, as the decay rate, so a small $|k_{\rm eff}|$ means a long range. Each curve is fitted only up to the lag where the mutual information first falls into the noise floor of the estimator. For display, the top row labels each target with its mean lag $\bar k$, the lag averaged with weights given by the mutual information, which orders the targets the same way.

\section{Effective Receptive Fields for \Trees}
\label{app:receptive-fields}

\begin{figure}[h]
\centering
\includegraphics[width=\textwidth]{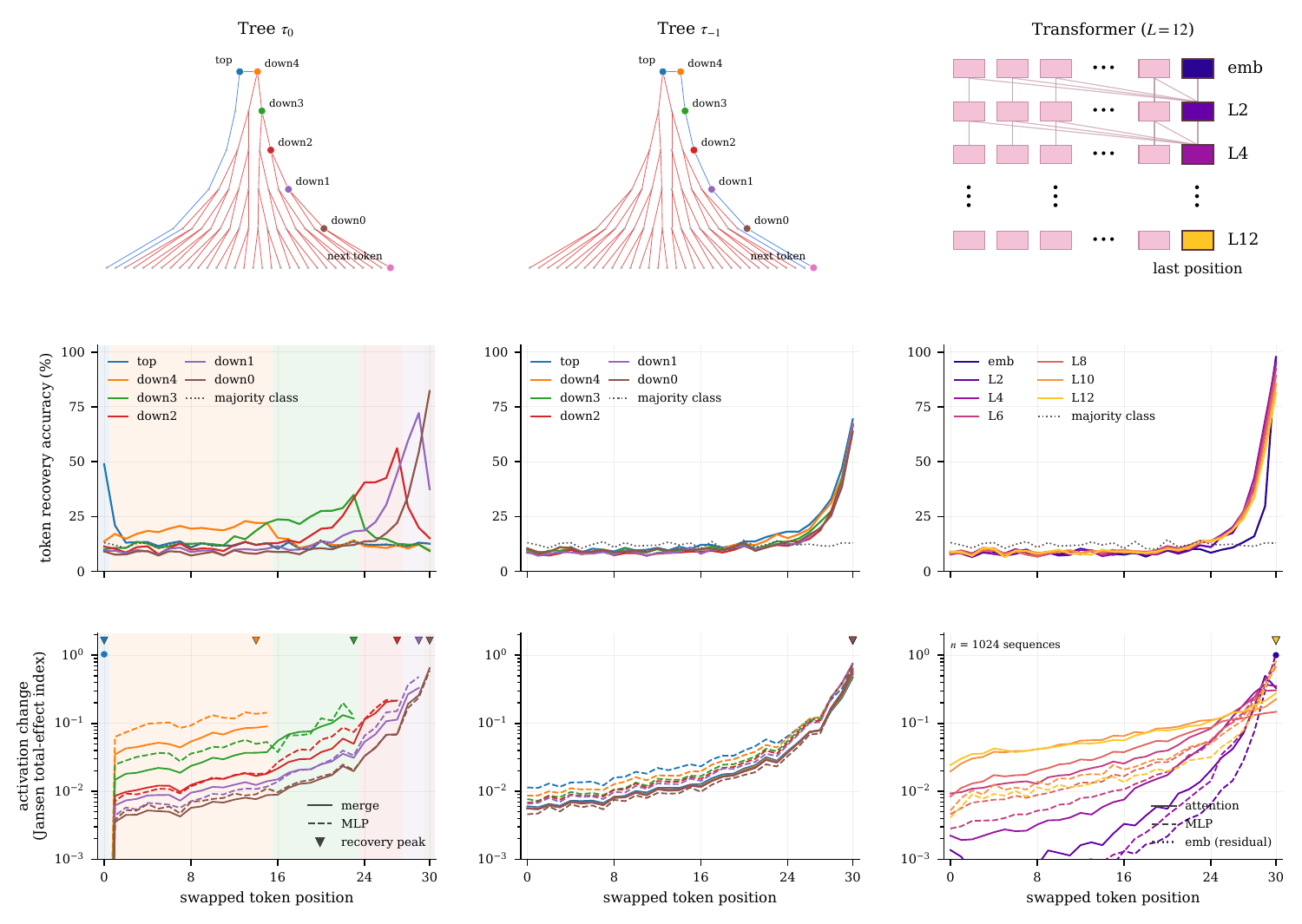}
\caption{Effective receptive fields. Columns: \tree with the aligned topology $\tau_0$, \tree with the anti-aligned topology $\tau_{-1}$, and a 12-layer transformer.
\textbf{Top:} the architecture, with the probed terminal nodes (trees) or last-position layers (transformer) marked in the colors used below; the tree leaves are aligned to the token axis.
\textbf{Middle:} we train linear probes to predict input tokens from latent activations of these nodes (top-1 accuracy; dotted: majority-class baseline). Accuracy for $\tau_0$ peaks at different intermediate context windows for different \tree layers, while the accuracy peaks at the last context position for all layers for $\tau_{-1}$ and transformers. This suggests that different layers in an aligned topology preserve information about different context windows, while the anti-aligned topology and transformers mix features for last-token prediction.
\textbf{Bottom:} receptive fields of individual nodes, recovered by activation patching. 
The token at each position is swapped, one position at a time, for the token another
validation sequence holds there (1024 sequences), and every activation is
compared with the unswapped run. We plot the mean squared change of each node's $\operatorname{Merge}$ (solid) and $\operatorname{MLP}$ (dashed) activation (attention and MLP for the transformer), normalized by twice the activation's variance (Jansen total-effect index, log scale). Outside a node's scope its activations are exactly
unchanged (no line).
Triangles mark where the linear probe of the middle row peaks; shaded bands mark the disjoint context windows of $\tau_0$.
For $\tau_0$, each node responds only to tokens up to the end of its window and most to the window it adds, peaking in the same windows as the probe; $x_0$ reaches the down nodes only below $10^{-8}$, outside the plotted range. In $\tau_{-1}$, no individual context windows are observed: every node's change rises smoothly towards the last token. The transformer behaves the same way at every layer.
}
\label{fig:patching}
\label{fig:patchnodes}
\label{fig:patchladder}
\end{figure}

\begin{figure}
    \centering
    \includegraphics[width=\linewidth]{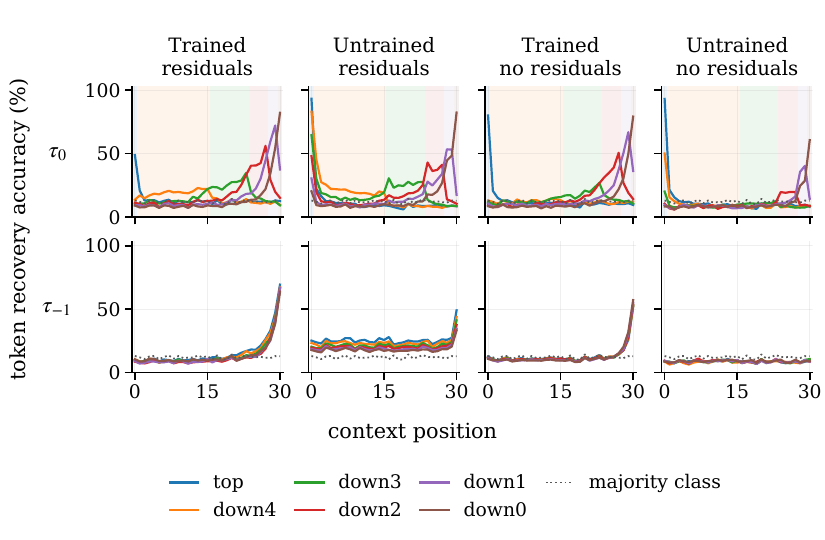}
    \caption{
    Do the effective receptive fields of $\tau_0$ come from training, from the topology, or from the residual connections? Linear-probe token recovery (as in Fig.~\ref{fig:patching}, middle row) for the six terminal nodes of $\tau_0$ (top row) and $\tau_{-1}$ (bottom row), in trained and untrained \trees with and without residual connections; dotted: majority-class baseline; shaded bands: the disjoint context windows of $\tau_0$. If the windows come from the wiring, they appear in the untrained $\tau_0$ but never in $\tau_{-1}$. They do: with residual connections, the untrained $\tau_0$ already recovers each window at its own node, and training sharpens the windows and removes the leakage of the first token into the lower nodes. Without residual connections, the untrained \tree loses tokens that pass through more than a few merges, and the trained one keeps only the narrow late windows. $\tau_{-1}$ peaks at the last token, or nowhere, in every condition.
    Panels show the mean of two seeds for untrained \trees with residual connections and trained \trees without them, and a single model otherwise. The trained $\tau_{-1}$ without residual connections is only partly trained (final losses of 3.44 and 2.85 nats for the two seeds).
    }
    \label{fig:trained-untrained}
\end{figure}

The aligned topology labelled $\tau_0$ (Figure~\ref{fig:ablation}) has the unique property that terminal nodes in different layers "focus" on different parts of the context. On the other hand, the anti-aligned topology $\tau_{-1}$ mixes information from the whole context.
We use a linear probe to uncover input tokens from latent activations of the terminal nodes in \trees. 
We observe that for $\tau_0$, different terminal nodes in different layers recover each input token in its scope.
The accuracy rises as we move down the layers, which indicates the difficulty of reconstruction increases as we pass through multiple non-linear passes.
In contrast, for $\tau_{-1}$ and transformers, recovery peaks at the last token, given the full context. 
This highlights that the aligned topology has \textit{effective receptive fields} for each node in the downward pass, in contrast with anti-aligned topologies and transformers.
Activation patching also confirms this causally (Fig.~\ref{fig:patching}). Restoring a single node recovers the prediction exactly when the swapped token lies in its scope and not at all otherwise.
We emphasize that we do not claim that transformers are implementing a $\tau_{-1}$ topology. Recovery of trees from transformers needs careful investigation due to known biasing effects from the parsing algorithm, feature definition as well as inherent architectural biases \citep{li2020branchingbiassyntaxextracted}.
In addition, we contrast this token recovery pattern in four different network ablations: trained and untrained \trees with or without residual connections (Fig.~\ref{fig:trained-untrained}). 
The windowed receptive fields of $\tau_0$ are \emph{already present} in an untrained \tree with residual connections: each terminal node recovers the tokens in its scope above the majority-class baseline at initialization, so the windows come from the wiring of the topology rather than from training. Training sharpens these windows and removes the leakage of the earliest tokens into the lower nodes.
Residual connections are what carry token identity through depth. Without them, an untrained \tree loses the tokens that pass through more than a few merges, and a trained one keeps the narrow windows near the end of the context but loses most of the wide early ones.
$\tau_{-1}$ shows no windows in \emph{any} of the four conditions: untrained, its nodes hold a weak, nearly uniform trace of the whole context, and trained, all of them peak at the last token.
We emphasize that the purpose of these experiments were to show that $\tau_0$ and $\tau_{-1}$ provide alternate mechanisms for model computation. The corresponding experiments with transformers show that this is distinct from the usual model computation, which are closer to the anti-aligned topology $\tau_{-1}$ as opposed to $\tau_0$.

\section{Goodness of fit for ranked path probabilities}
\label{app:routing-gof}

We sort the exact path probabilities and normalize each candidate on the finite rank support, $r=1,\ldots,36720$. The probability of a path is the token average of its per-token joint probability, $p(\mathrm{path})=\mathbb{E}_x\big[\prod_{\ell}p^{(\ell)}(g_\ell\mid x)\big]$, over $192{,}448$ validation tokens.We fit each router and seed separately, by least squares on $\log p(r)$. This gives equal weight to each rank and resolves the low-probability tail. The primary measures are log-RMSE and $\Delta$AICc. AICc assumes Gaussian log residuals. KS and $D_{\mathrm{KL}}(p\Vert q)$ are secondary descriptive checks.

\begin{table}[h!]
\centering
\caption{Goodness of fit for the ranked path-probability curves, fitted separately to the two seeds (s0, s1) of each router. Lower is better in every column, and the best value in each column of a block is in bold. PL denotes power law. GEC denotes generalized exponential cutoff. DGBD denotes the discrete generalized Beta distribution discussed in \citep{naumis2007tails}.}
\label{tab:routing-gof}
\scriptsize
\setlength{\tabcolsep}{2.2pt}
\resizebox{\textwidth}{!}{%
\begin{tabular}{lrrrrrrrr}
\toprule
& \multicolumn{2}{c}{log-RMSE} & \multicolumn{2}{c}{$\Delta$AICc} & \multicolumn{2}{c}{KS} & \multicolumn{2}{c}{KL} \\
\cmidrule(lr){2-3}\cmidrule(lr){4-5}\cmidrule(lr){6-7}\cmidrule(lr){8-9}
Model & s0 & s1 & s0 & s1 & s0 & s1 & s0 & s1 \\
\midrule
\multicolumn{9}{l}{$m=1$} \\
Exponential                & 2.965 & 2.972 & 191424.8 & 175489.6 & 0.839 & 0.809 & 3.597 & 3.513 \\
Generalized exponential    & 0.433 & 0.393 &  50186.3 &  26910.6 & 0.373 & 0.340 & 0.580 & 0.590 \\
Power law / Pareto         & 1.622 & 1.940 & 147131.1 & 144180.5 & 0.431 & 0.459 & 0.682 & 0.880 \\
PL with exponential cutoff & 0.351 & 0.371 &  34644.7 &  22679.1 & 0.266 & 0.279 & 0.200 & 0.259 \\
PL with GEC                & \textbf{0.219} & \textbf{0.272} & \textbf{0.0} & \textbf{0.0} & \textbf{0.087} & \textbf{0.120} & \textbf{0.068} & \textbf{0.128} \\
Broken PL / double Pareto  & 0.281 & 0.467 &  18371.9 &  39542.0 & 0.271 & 0.286 & 0.211 & 0.272 \\
DGBD / beta-rank           & 0.939 & 1.007 & 106993.4 &  96049.9 & 0.375 & 0.398 & 0.443 & 0.559 \\
\midrule
\multicolumn{9}{l}{$m=2$} \\
Exponential                & 1.736 & 2.288 & 179757.3 & 192998.3 & 0.743 & 0.879 & 2.797 & 4.086 \\
Generalized exponential    & 0.369 & 0.735 &  66086.3 & 109589.8 & 0.341 & 0.565 & 0.417 & 0.956 \\
Power law / Pareto         & 0.852 & 0.554 & 127501.0 &  88814.8 & 0.317 & 0.244 & 0.333 & 0.161 \\
PL with exponential cutoff & 0.208 & 0.255 &  24092.6 &  31735.5 & 0.127 & 0.155 & 0.066 & 0.090 \\
PL with GEC                & 0.201 & 0.242 &  21497.1 &  27930.4 & \textbf{0.095} & \textbf{0.120} & \textbf{0.044} & \textbf{0.074} \\
Broken PL / double Pareto  & \textbf{0.150} & \textbf{0.165} & \textbf{0.0} & \textbf{0.0} & 0.178 & 0.154 & 0.114 & 0.090 \\
DGBD / beta-rank           & 0.512 & 0.392 &  90046.1 &  63404.0 & 0.241 & 0.209 & 0.193 & 0.125 \\
\bottomrule
\end{tabular}}
\end{table}

For $m=1$ the power law with a generalized exponential cutoff is best on every measure for both seeds, with log-scale $R^2=0.9959$ and $0.9948$. For $m=2$ the broken power law has the lowest log-RMSE and AICc for both seeds ($R^2=0.9953$ and $0.9929$), while the generalized cutoff has the lowest KS and KL. Both place the bend at $r\approx6{,}000$--$9{,}000$, one as a sharp break and one smoothly. Under the generalized cutoff, $p(r)\propto r^{-\alpha}e^{-(r/r_c)^{\beta}}$, the cutoff scale separates the two routers, $r_c=541$ and $777$ for $m=1$ against $6{,}860$ and $8{,}170$ for $m=2$, while $\alpha$ ($1.16$--$1.33$) and $\beta$ ($0.53$--$0.81$) vary about as much between the two seeds of one router as between routers. The $100$ most probable paths carry $88\%$ and $85\%$ of the mass for $m=1$ and $70\%$ and $91\%$ for $m=2$, but their identity does not reproduce: the two seeds of a router share none of their $100$ most probable paths, and only $3$ ($m=1$) and $0$ ($m=2$) of their top $500$, against about $7$ expected by chance. Pareto and power law are the same finite rank model here. Double Pareto and sharp broken power law are also the same rank curve. The smooth cutoff agrees with the transition from power-law-like ranks to faster tails discussed by \citep{naumis2007tails}.

\clearpage
\newpage
\section{Examples of generated text}
\label{app:gen_data}

Samples from fixed \trees on TinyStories (BPE-$8192$) at temperature $1$, with a fixed context and generation seed. Within each block every model receives the same validation prompt, truncated to its own context length $n-1$; the token stream is lightly de-tokenised (punctuation re-attached) for readability, and \textsc{[eos]} marks the end-of-story token.

\subsection*{I. Width sweep --- $d$ varies at $n=32$}

\paragraph{$n = 32$, $d = 64$ \quad($31$ context tokens, $h_{\mathrm{pred}} = 2.537$ nats/token)}\mbox{}\\

\textit{Prompt.}~camp. \textsc{[eos]} One day, a little girl named Lucy found a shoe in her room. It was a big shoe, much too big for her foot.\\

\textit{Continuation.}~Mia tried to climb up, but she couldn't. She felt so cold and strong. She did not shiver anymore. Her music made her. It made her bow back on the ground. The lemonade tasted like Mr. H ina's face turned into a small sc run. The fit and introduced all her was out to dry and decided to show it to her, making a pile castle. She had some money from

\paragraph{$n = 32$, $d = 128$ \quad($31$ context tokens, $h_{\mathrm{pred}} = 2.247$ nats/token)}\mbox{}\\

\textit{Prompt.}~camp. \textsc{[eos]} One day, a little girl named Lucy found a shoe in her room. It was a big shoe, much too big for her foot.\\

\textit{Continuation.}~Mia wanted to comfort Fluffy, but she was scared. She did not know what to play with. Kitty's mom saw her feeling sad and did not want to help. So, she hid the medicine in the kitchen. \textsc{[eos]} One day, a colorful bird named Bob flew in a big tree. March was a happy place to be. He was very happy to find the sun. Bobo flew to a big park where people

\paragraph{$n = 32$, $d = 256$ \quad($31$ context tokens, $h_{\mathrm{pred}} = 1.938$ nats/token)}\mbox{}\\

\textit{Prompt.}~camp. \textsc{[eos]} One day, a little girl named Lucy found a shoe in her room. It was a big shoe, much too big for her foot.\\

\textit{Continuation.}~Lucy tried to fix the shoe, but it couldn't pinch her dress. Amy asked her mom, ``Mom, what is this jacket?'' Her mom said, ``I don't know, Sue. Maybe they can use a glove to talk to someone.'' Finally, Tim's mom mentioned the coin to Tim. But when they went back home, Tim forgot about Spot. She gave it back to Tim and said

\paragraph{$n = 32$, $d = 512$ \quad($31$ context tokens, $h_{\mathrm{pred}} = 1.890$ nats/token)}\mbox{}\\

\textit{Prompt.}~camp. \textsc{[eos]} One day, a little girl named Lucy found a shoe in her room. It was a big shoe, much too big for her foot.\\

\textit{Continuation.}~She tried to pull it, but she couldn't. She felt frustrated. Amy asked her mom, ``Mom, what is this round thing?'' Her mom smiled and said, ``This is a special seat, Amy. It tells us things A test is when we drive around the front to the whole world. We can learn this and learn how to count to 10, to the beauty end. How does that sound?''

\paragraph{$n = 32$, $d = 1024$ \quad($31$ context tokens, $h_{\mathrm{pred}} = 1.756$ nats/token)}\mbox{}\\

\textit{Prompt.}~camp. \textsc{[eos]} One day, a little girl named Lucy found a shoe in her room. It was a big shoe, much too big for her foot.\\

\textit{Continuation.}~She wanted to repair it, but she did not know how. ``Mom, can I reverse our shoe?'' Sam requested. His mom smiled and said, ``Of course we can! I'll hold it for you.'' \textsc{[eos]} One day, a colorful bird named Bob flew to a big tree. He wanted to settle there to rest and rest. While he was sitting, he heard a loud laugh. Rob opened his eyes

\subsection*{II. Context sweep --- $n$ varies at $d=1024$}

\paragraph{$n = 3$, $d = 1024$ \quad($2$ context tokens, $h_{\mathrm{pred}} = 2.843$ nats/token)}\mbox{}\\

\textit{Prompt.}~too big\\

\textit{Continuation.}~for you. Do you want to buy the blue ball. One sunny day, a boy named Tom came to the war and be friends and her white dress. She ran back home. They played together and had fun. Would you like.``Bob wanted to bake with her toys and games. They agreed and they both loved the doll.'' You can't need to sleep. \textsc{[eos]} Once upon a time, there was a

\paragraph{$n = 4$, $d = 1024$ \quad($3$ context tokens, $h_{\mathrm{pred}} = 2.537$ nats/token)}\mbox{}\\

\textit{Prompt.}~much too big\\

\textit{Continuation.}~for you. Do you want to buy the blue ball. One sunny day, a boy named Tom came to Sam's orange collar and her white boots. She ran back home to play together. \textsc{[eos]} Once upon a time, there was a rabbit. She had seen the wind's joy and introduced himself. The cat was busy and decided to show it to all her friends. She gave Lily a big hug and said

\paragraph{$n = 8$, $d = 1024$ \quad($7$ context tokens, $h_{\mathrm{pred}} = 2.085$ nats/token)}\mbox{}\\

\textit{Prompt.}~a big shoe, much too big\\

\textit{Continuation.}~and old. John felt frustrated. He yelled,'But the parrot wouldn't listen. He came closer and saw what was happening. She did not want to be rude. Tom looked at Sam and said, ``Would you like some more?'' But Tom could not hear Tim's voice and shake the hole. A big dog came to the castle. The dog did not like the decorations. He wanted some money from

\paragraph{$n = 16$, $d = 1024$ \quad($15$ context tokens, $h_{\mathrm{pred}} = 1.818$ nats/token)}\mbox{}\\

\textit{Prompt.}~in her room. It was a big shoe, much too big for her\\

\textit{Continuation.}~mouth. Sue didn't want her spaghetti, but her mom said, ``Don't worry, we can make another one later.'' After that, they played a game where Tom squeezed the bowl and made a loud noise. ``Boo!'' A sound came out. ``Wow, you fit us! You are good at hiding. You can escape through the hole to sleep.'' Tim followed Max to a big park where there

\paragraph{$n = 32$, $d = 1024$ \quad($31$ context tokens, $h_{\mathrm{pred}} = 1.756$ nats/token)}\mbox{}\\

\textit{Prompt.}~camp. \textsc{[eos]} One day, a little girl named Lucy found a shoe in her room. It was a big shoe, much too big for her foot.\\

\textit{Continuation.}~She wanted to repair it, but she did not know how. ``Mom, can I reverse our shoe?'' Sam requested. His mom smiled and said, ``Of course we can! I'll hold it for you.'' \textsc{[eos]} One day, a colorful bird named Bob flew to a big tree. He wanted to settle there to rest and rest. While he was sitting, he heard a loud laugh. Rob opened his eyes

\end{document}

%% file: math_commands.tex
\usepackage{amsmath,amsfonts,bm}

\def\eqref#1{equation~\ref{#1}}

\def\1{\bm{1}}

\DeclareMathAlphabet{\mathsfit}{\encodingdefault}{\sfdefault}{m}{sl}
\SetMathAlphabet{\mathsfit}{bold}{\encodingdefault}{\sfdefault}{bx}{n}



%% file: appendix_dyadic_toy_model.tex
\label{app:dyadic-toy}

We provide details of the toy model that we defined in the main text, which suggests a plausible mechanism for the HP transfer experiment results we observe empirically.

\begin{figure}
    \centering
    \includegraphics[width=\linewidth]{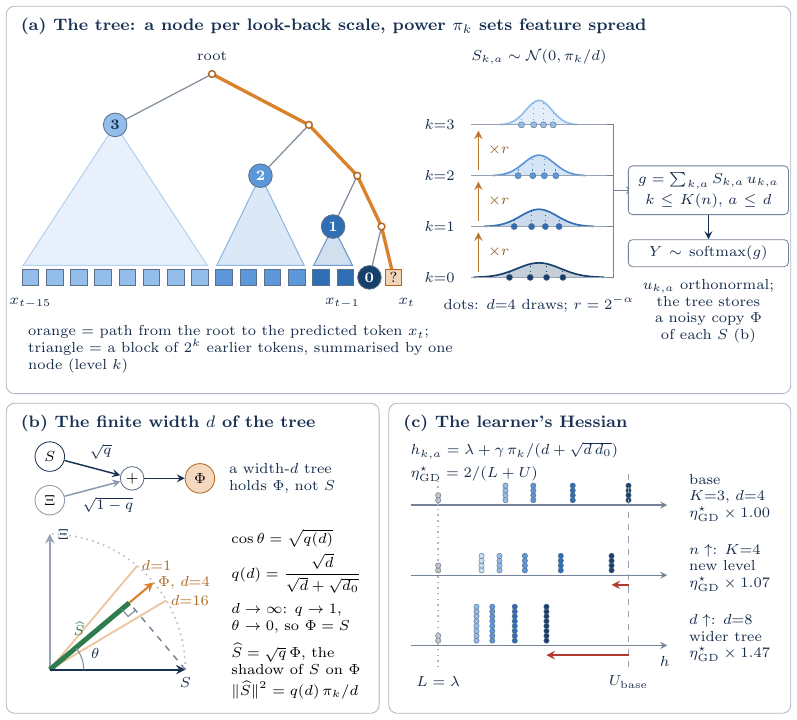}
    \caption{Why does the optimal step size drift with the context length $n$ and the
channel count $d$? The figure follows one next-token example through the toy
model and then reads off the curvature that a quadratic learner would see.
\textbf{(a)}~Each doubling of the look-back window reveals one new block of
context (colour), and each block becomes one level $k$ that carries $d$
independent channels. Each channel has variance $\pi_k/d$,
shown by the Gaussian spread. The level powers decay geometrically
and sum to one. Adding a level redistributes this fixed budget.
The label depends on the clean features $S$ alone; the learner
sees only the noisy copy $\Phi$. No arrow joins different levels or channels,
so each channel is estimated on its own. Drawn for $\alpha=1/2$, $K=3$, $d=4$.
\textbf{(b)}~One channel in rescaled coordinates, with equal fixed
lengths for $S$ and $\Phi$. Their angle satisfies $\cos^2\theta=q(d)$.
The projection is $\widehat S=\sqrt q\,\Phi$, with power
$\pi_k/(d+\sqrt{d\,d_0})$. \textbf{(c)}~The
Hessian eigenvalue of each channel, stacked by multiplicity, for $\alpha=1/2$,
$d_0=2$, $\lambda=0.01$, $\gamma=1$ (grey: ridge-only directions). Under the toy
model, adding a level or adding channels moves the stiffest eigenvalue $U$ to
the left, and $\eta^\star_{\rm GD}$ rises; the shift from $d$ is much larger than
the shift from $n$.}
\label{fig:dy-intuition}
\end{figure}

This model is based primarily on the random hierarchy model \citep{PhysRevX.14.031001} and the wavelet analysis of \citep{wavelet-model, mallat1989theory}. The toy model for the data has two layers.
A finite dyadic hierarchy allocates predictive power across look-back scales. 
A Gaussian channel limits how well each latent feature can be recovered. 
Each of these layers are used to implement a different mode of control that \trees offer. We discuss this in more detail below.
We make an attempt to highlight every assumption that a conclusion depends on through the labels (A1) -- (A10).
Figure~\ref{fig:dy-intuition} gives an intuitive picture.

\subsection{Scales and signal strength}
\label{app:dy-scales}

Let $X=(x_{t-n+1},\ldots,x_{t-1})$ be a context of $n-1$ tokens, and let
$g(X)=\log p(\cdot\mid X)-V^{-1}\sum_y\log p(y\mid X)\,\mathbf 1$ be its
centered Bayes logit vector. We assume $p(y\mid X)>0$ and
$\mathbb E\|g\|^2<\infty$. The dyadic suffix window $W_k$ holds the $2^k$ most
recent tokens, so $W_0\subset W_1\subset\cdots$. Passing from $W_{k-1}$ to
$W_k$ reveals the tokens at distances $2^{k-1}+1,\ldots,2^k$; for $k=0$, it
reveals $x_{t-1}$. This is the coarse-to-fine organization of multiresolution
analysis~\citep{mallat1989theory}. Let $\mathcal W_k$ be the $\sigma$-algebra
generated by $W_k$, and let $\mathcal W_{-1}$ be the trivial one. The level-$k$
increment and its signal strength are
\begin{equation}
\Delta_k=\mathbb E[g\mid\mathcal W_k]-\mathbb E[g\mid\mathcal W_{k-1}],
\qquad
s_k=\mathbb E\big[\Delta_k^\top F_0\,\Delta_k\big],
\label{eq:dy-increment}
\end{equation}
where $F_0=\operatorname{diag}(p_0)-p_0p_0^\top$ is the categorical Fisher
matrix at the uniform distribution $p_0$. It is positive definite on centered
vectors, and we write $\|v\|_{F_0}^2=v^\top F_0v$. The increments are
martingale differences. For a constant metric they are therefore orthogonal,
$\mathbb E[\Delta_j^\top F_0\Delta_k]=0$ for $j\neq k$, and their powers add:
$\sum_{k\le K}s_k=\mathbb E\|\mathbb E[g\mid\mathcal W_K]-\mathbb E[g]\|_{F_0}^2$.
The strength $s_k$ is signal power, not a Hessian eigenvalue.

\medskip\noindent\textbf{(A1) Modelling the effect of horizon  and capacity equivalents} 
The horizon $n$ and the channel count $d$ are hyperparameters of the data distribution. In our experiments, they stand for the context window and the width of the \tree that generated a corpus. 
Here, we attempt to model the dependence of the distribution on these hyper-parameters through the toy model that we describe above.
$n$ sets a finite context window by an abrupt cutoff on the lookback distance
$K(n)=\lfloor\log_2(n-1)\rfloor$.
$d$ controls the fidelity of the different modes through a quality factor $q(d)$ which we define later.

\medskip\noindent\textbf{(A2) Suffix windows, one summary per level.} The model keeps one $d$-channel summary per level and ignores the internal organization of each new block.

\subsection{Scale powers}
\label{app:dy-powers}

A hierarchy alone does not determine $s_k$. Suppose that the level-$k$
increment has amplitude $C_k=C_0\prod_{\ell=1}^{k}A_\ell$, with
$s_k=\mathbb E[C_k^2]$.

\medskip\noindent\textbf{(A3) Self-similar branch law.} The factors
$C_0,A_1,A_2,\ldots$ are mutually independent, and
$\mathbb E[A_\ell^2]=2^{-\alpha}$ for every $\ell$.

\medskip\noindent
Then $s_k=\mathbb E[C_0^2]\,2^{-\alpha k}$. Such a power law was also obtained in \citep{lin2017critical}. 
The empirically relevant case is $\alpha>0$. Only the levels whose window fits in the context are
active:
\begin{equation}
K(n)=\lfloor\log_2(n-1)\rfloor .
\label{eq:dy-cutoff}
\end{equation}
This sets the context approximation in \medskip\noindent \textbf{(A1)}. Hence $K$ is constant for $n\in\{2^K+1,\ldots,2^{K+1}\}$, and the tokens at
distances $2^{K(n)}+1,\ldots,n-1$ are dropped.

\medskip\noindent\textbf{(A4) Fixed total budget.} For every $n$, the total
clean power is one: $\sum_ks_k=1$.

\medskip\noindent
With $r=2^{-\alpha}$, the level powers are
\begin{equation}
\pi_k(n)=\frac{r^k}{Z_{K(n)}}\,\mathbf 1[k\le K(n)],
\qquad
Z_K=\sum_{\ell=0}^{K}r^\ell=
\begin{cases}
(1-r^{K+1})/(1-r), & r\neq1,\\
K+1, & r=1.
\end{cases}
\label{eq:dy-pi}
\end{equation}
The largest level power is $\pi_{\max}(K)=1/Z_K$ for $\alpha\ge0$ and
$r^K/Z_K=1/\sum_{j=0}^{K}r^{-j}$ for $\alpha<0$. It decreases strictly with $K$
and tends to $1-r$, $0$, and $1-1/r$ for $\alpha>0$, $\alpha=0$, and
$\alpha<0$. Assumption~(A4) deserves care. For one fixed source, the
increments~\ref{eq:dy-increment} do not depend on $n$, so a longer context
adds levels without weakening the old ones. The normalization therefore
describes a family of teachers, one for each $n$, each scaled to unit
predictable power. It is also the only source of $n$-dependence in the
spectral edge below. Without it, $\max_kr^k$ is independent of $n$ for
$\alpha\ge0$ and grows with $n$ for $\alpha<0$.

\subsection{The feature--label generator}
\label{app:dy-generator}

For each sample, each active level $k$, and each channel $a=1,\ldots,d$, draw
\begin{equation}
S_{k,a},\ \Xi_{k,a}\ \overset{\rm iid}{\sim}\ \mathcal N\big(0,\pi_k(n)/d\big),
\qquad
\Phi_{k,a}=\sqrt{q}\,S_{k,a}+\sqrt{1-q}\,\Xi_{k,a},
\qquad
q(d)=\frac{\sqrt{d}}{\sqrt{d}+\sqrt{d_0}}.
\label{eq:dy-channel}
\end{equation}
Choose centered logit directions $u_{k,a}\in\mathbb R^V$ that are orthonormal
in the $F_0$ metric, which requires $d\,(K(n)+1)\le V-1$. Generate
\begin{equation}
g(S)=\sum_{k=0}^{K(n)}\sum_{a=1}^{d}S_{k,a}\,u_{k,a},
\qquad
Y\mid S\sim\operatorname{Categorical}\big(\operatorname{softmax}(g(S))\big).
\label{eq:dy-label}
\end{equation}
The dataset contains $(\Phi,Y)$, not $S$ or $\Xi$. The index $k$ labels the
level and $a$ labels its channels. Each channel has its own direction
$u_{k,a}$, so the $d$ features of a level stay distinguishable in their effect
on the prediction, and $d$ is the number of resolved features per level. The
latent summaries play the role of the context in
Section~\ref{app:dy-scales}: $\Delta_k=\sum_aS_{k,a}u_{k,a}$ and
$s_k=\pi_k(n)$. The base logit is zero, so $p_0$ is uniform. A non-uniform
unigram distribution would enter as a fixed logit offset.

\medskip\noindent\textbf{(A5) Independent, balanced Gaussian channels.} All $S_{k,a}$ and $\Xi_{k,a}$ are jointly Gaussian and independent, and the $d$ channels of a level share its power equally.

\medskip\noindent\textbf{(A6) Scale-uniform recovery.}
The recovery fraction $q(d)=\sqrt{d}/(\sqrt{d}+\sqrt{d_0})$ is the same at every level and does not depend on $n$. Its deficit satisfies $1-q(d)\sim\sqrt{d_0/d}$ as $d\to\infty$.

\medskip\noindent
The per-channel signal-to-noise ratio is $q/(1-q)=\sqrt{d/d_0}$.
Because the channel is Gaussian, $S_{k,a}\mid\Phi_{k,a}\sim\mathcal N\big(\sqrt q\,\Phi_{k,a},(1-q)\pi_k/d\big)$,
and by independence $\mathbb E[S_{k,a}\mid\Phi]=\mathbb E[S_{k,a}\mid\Phi_{k,a}]$.
The posterior mean and its power are
\begin{equation}
\widehat S_{k,a}=\sqrt q\,\Phi_{k,a},
\qquad
\omega_k(n,d)\equiv\operatorname{Var}(\widehat S_{k,a})
=q(d)\,\frac{\pi_k(n)}{d}
=\frac{\pi_k(n)}{d+\sqrt{d\,d_0}}.
\label{eq:dy-recoverable}
\end{equation}
The recoverable covariance is
$\Sigma_{\rm rec}=\operatorname{diag}\{\omega_k\}=q(d)\,\Sigma_\Phi$, with
trace $q(d)$. More capacity recovers more total signal, but each channel
carries less of it: $\omega_k$ falls with $d$ while
$\operatorname{tr}\Sigma_{\rm rec}$ rises. The total recoverable power
$\operatorname{tr}\Sigma_{\rm rec}=q(d)$ tends to zero as $d\to0$.
The normalized spectrum
$\omega_k/\operatorname{tr}\Sigma_{\rm rec}=\pi_k/d$
is known in closed form, and no corpus statistic enters it.

\subsection{A matched quadratic learner}
\label{app:dy-learner}

We now provide a minimal quadratic model to set up the learner for the data distribution above.

The learner has one coordinate $\theta_{k,a}$ for every channel it can
represent, including channels that are inactive at the current $(n,d)$. It
predicts $f_\theta(\Phi)=\sqrt\gamma\sum_{k,a}\theta_{k,a}\widehat
S_{k,a}u_{k,a}$. Take a fixed regression target $g^\dagger(\Phi)$, such as
$\mathbb E[g(S)\mid\Phi]=\sum_{k,a}\widehat S_{k,a}u_{k,a}$. The population
objective and its Hessian are
\begin{equation}
\mathcal Q(\theta)
=\tfrac12\,\mathbb E\,\|f_\theta(\Phi)-g^\dagger(\Phi)\|_{F_0}^2
+\tfrac\lambda2\|\theta\|_2^2,
\qquad
H=\lambda I+\gamma\Sigma_{\rm rec},
\qquad
h_{k,a}=\lambda+\gamma\,\omega_k(n,d).
\label{eq:dy-hessian}
\end{equation}
The off-diagonal entries
$\gamma\,\mathbb E[\widehat S_{k,a}\widehat S_{k',a'}]\,u_{k,a}^\top F_0u_{k',a'}$
vanish because the output directions are $F_0$-orthonormal, even if the
features are correlated. Inactive coordinates have $\widehat S\equiv0$ and
curvature $\lambda$. Data features and optimization modes therefore align by
construction. The target moves the optimum but not the Hessian.

\medskip\noindent\textbf{(A7) Matched features.} The learner reads $\widehat S=\sqrt q\,\Phi$, not $\Phi$. This is a strong assumption. We justify it here by assuming that the warmup phase of learning aligns the model's features with the data features, so that the optimal learning rate is driven in this pseudo-constant phase where the learner model has aligned features.

\medskip\noindent\textbf{(A8) Weak signal.} Because $f_\theta$ is linear in
$\theta$, the regularized cross-entropy objective has the exact Hessian
$H_{\rm CE}(\theta)=\mathbb E[J^\top F(p_\theta(\cdot\mid\Phi))J]+\lambda I$,
with $J=\partial f_\theta/\partial\theta$ and
$p_\theta=\operatorname{softmax}(f_\theta)$. Equation~\ref{eq:dy-hessian}
replaces $F(p_\theta)$ by $F_0$. This assumes that predictions stay near the
reference distribution, that is, that the total clean power $\sigma^2$ is small
in $F_0$ units. Assumption~(A4) sets $\sigma^2=1$, and a smaller $\sigma^2$ is
absorbed into $\gamma$.

\medskip\noindent
The replacement is exact at $\theta=0$. At any $\theta$, the correction that is
first order in the logits vanishes in expectation, because it is odd under
$(S,\Xi)\to(-S,-\Xi)$. The error is therefore of relative order $\sigma^2$.
Numerically, use the square-root recovery law
$q(d)=\sqrt{d}/(\sqrt{d}+\sqrt{d_0})$ with $V=16$, $K=2$, $d=3$,
$d_0=2$, $r=1/2$, $\gamma=1$, and $\lambda=10^{-3}$.
For each of $24$ random draws of the $u_{k,a}$, we use $60{,}000$
Gaussian feature draws and minimize the sample-average cross-entropy
plus the ridge penalty.
At the numerical optimum, the median of
$\|H_{\rm CE}-H\|_{\rm F}/\|H\|_{\rm F}$ is
$1.00\%$, $10.61\%$, and $25.98\%$ at
$\sigma^2=0.1$, $1$, and $4$, respectively.
Both Hessians are evaluated on the same samples.
The unit budget corresponds to about half a nat of divergence from
the uniform distribution, since
$\mathrm{KL}(\operatorname{softmax}(g)\,\|\,p_0)\approx\frac12\|g\|_{F_0}^2$.
The quadratic model is exact for a Gaussian regression target
$\widetilde Y=g(S)+\varepsilon$ with independent centered noise, since then
$\frac12\mathbb E\|f_\theta-\widetilde Y\|_{F_0}^2
=\frac12\mathbb E\|f_\theta-\mathbb E[g\mid\Phi]\|_{F_0}^2+\text{const}$.

\subsection{Spectral edges and the optimal learning rate}
\label{app:dy-rate}

\medskip\noindent\textbf{(A9) Ridge-only directions.} At least one learner
coordinate has no data curvature, for example an inactive channel, so
$L=\lambda$. Otherwise, $L=\lambda+\gamma\pi_{\min}/(d+\sqrt{d\,d_0})$ also moves with
$(n,d)$.

\medskip\noindent\textbf{(A10) Deterministic, non-adaptive optimizer.} The
optimizer takes full-batch steps on $\mathcal Q$ with constant parameters.
Minibatch noise, schedules, and Adam-type preconditioning are outside the
model.

\medskip\noindent
The upper edge is
\begin{equation}
U(n,d)=\lambda+\gamma\,\frac{\pi_{\max}(K(n))}{d+\sqrt{d\,d_0}}.
\label{eq:dy-upper}
\end{equation}
For a spectrum in $[L,U]$, gradient descent
$\theta_{t+1}=\theta_t-\eta\nabla\mathcal Q(\theta_t)$ has the minimax step
$\eta^\star_{\rm GD}=2/(L+U)$. This step equalizes $1-\eta L=\eta U-1$ and
contracts at rate $(U-L)/(U+L)$. The heavy-ball update
$\theta_{t+1}=\theta_t-\eta\nabla\mathcal Q(\theta_t)+\beta(\theta_t-\theta_{t-1})$
has $\eta^\star_{\rm HB}=4/(\sqrt L+\sqrt U)^2$ and
$\beta^\star_{\rm HB}=\big((\sqrt U-\sqrt L)/(\sqrt U+\sqrt L)\big)^2$
\citep{polyak1964some,danilova2022averaged}. Nesterov's method,
$y_t=\theta_t+\beta(\theta_t-\theta_{t-1})$ and
$\theta_{t+1}=y_t-\eta\nabla\mathcal Q(y_t)$, commonly uses $\eta=1/U$ and
$\beta=(\sqrt U-\sqrt L)/(\sqrt U+\sqrt L)$~\citep{assran2020convergence}. With
$L=\lambda$, the gradient-descent rate decomposes exactly as
\begin{equation}
\nu^\star_{\rm GD}(n,d)\equiv\log_2\eta^\star_{\rm GD}
=\underbrace{\log_2\frac1\lambda}_{\nu^\star_\infty}
\;-\;\underbrace{\log_2\!\Big(1+\frac{\gamma\,\pi_{\max}(K(n))}{2\lambda\,(d+\sqrt{d\,d_0})}\Big)}_{\text{capacity term}\ \ge\ 0}.
\label{eq:dy-nu}
\end{equation}
The capacity term decreases strictly in $d$ and stepwise in $n$.
It vanishes as $d\to\infty$ and saturates in $n$ for $\alpha\neq0$.
When $U\gg L$ and $\alpha\ge0$, Eq.~\ref{eq:dy-nu} becomes
$\nu^\star_{\rm GD}\simeq
1-\log_2\gamma+\log_2(d+\sqrt{d\,d_0})+\log_2Z_{K(n)}$.
More generally, $\nu^\star_A=C_A-\log_2U+o(1)$, with $C_{\rm GD}=1$, $C_{\rm HB}=2$, and $C_{\rm NAG}=0$. The heavy-ball correction is only $O(\sqrt{L/U})$. 
The stability threshold $2/U$ has the same scaling, so the prediction concerns the dependence on $U$ rather than the optimality criterion.

The monotone drift uses (A1), (A4), (A5), and (A9), together with two facts:
new levels receive positive power, and recovery grows sublinearly, so that
$q(d)/d=1/(d+\sqrt{d\,d_0})$ decreases and dilution beats recovery. The drift can fail if
the total power grows with depth, if recovery improves faster than linearly over
some range of $d$, or if an unrelated direction sets $U$.

\subsection{From the solvable learner to a transformer}
\label{app:dy-alignment}

Equation~\ref{eq:dy-hessian} makes the data features and the Hessian
eigenvectors coincide by design. Applying Eq.~\ref{eq:dy-nu} to a transformer
requires a hypothesis in three parts. \textbf{(H1)} The dominant Hessian
eigendirections align approximately with the multiscale predictive features.
\textbf{(H2)} The top eigenvalue belongs to such a direction and equals an
$(n,d)$-independent gain times its recoverable power. \textbf{(H3)} The lower
spectral edge does not depend on $(n,d)$.

For a well-specified softmax model at a population optimum, the cross-entropy
Hessian reduces to its Gauss--Newton part,
\begin{equation}
H_{\rm CE}=\mathbb E_X\big[J_\theta^\top F(p_\theta(\cdot\mid X))\,J_\theta\big],
\label{eq:dy-gauss-newton}
\end{equation}
because the residual term
$\mathbb E_X\big[\sum_y(p_\theta(y\mid X)-p(y\mid X))\nabla^2_\theta
f_{\theta,y}(X)\big]$ vanishes there. The strengths $s_k$ determine neither the
singular vectors nor the singular values of the learned logit Jacobian
$J_\theta$. A transformer can rotate the features, mix scales, and add
unrelated high-curvature directions. The Fisher factor adds a corpus dependence
of its own: $F(p_\theta)$ shrinks as predictions sharpen, so the gain in (H2)
can itself depend on $(n,d)$. The learned Jacobian is the unresolved
first-principles bottleneck.